\documentclass{article}
\usepackage{arxiv}
\usepackage[utf8]{inputenc}
\usepackage[T1]{fontenc}
\usepackage{booktabs}
\usepackage{amsfonts}
\usepackage{nicefrac}
\usepackage{microtype}
\usepackage{lipsum}
\usepackage{graphicx}
\usepackage[backend=biber,style=numeric,maxnames=6,minnames=1,sorting=none]{biblatex}
\usepackage{doi}
\usepackage{multirow}
\usepackage{multicol}
\usepackage{float}
\usepackage{enumerate}
\usepackage{threeparttable}
\usepackage{authblk}
\usepackage{CJKutf8}
\usepackage{changes}
\usepackage{amsmath}
\usepackage{subcaption}
\usepackage{caption}
\newcommand{\bench}{AECBench}

\usepackage{caption}
\title{\bench{}: A Hierarchical Benchmark for Knowledge Evaluation of Large Language Models in the AEC Field}

\author[1,3]{Chen Liang}
\author[1]{Zhaoqi Huang}
\author[4]{Haofen Wang}
\author[1,2]{Fu Chai}
\author[1]{Chunying Yu}
\author[2]{Huanhuan Wei}
\author[1,2]{Zhengjie Liu}
\author[2]{Yanpeng Li}
\author[2]{Hongjun Wang}
\author[1,2]{Ruifeng Luo}
\author[1,3]{Xianzhong Zhao}
\affil[1]{College of Civil Engineering, Tongji University, 1239 Siping Road, Shanghai 200092, China}
\affil[2]{Arcplus Group East China Architectural Design \& Research Institute Co., Ltd., 151 Hankou Road, Shanghai 200002, China}
\affil[3]{Shanghai Qi Zhi Institute, 701 Yunjin Road, Shanghai 200232, China}
\affil[4]{College of Design and Innovation, Tongji University, 1239 Siping Road, Shanghai 200092, China}
\date{}
\begin{document}
\maketitle
\footnotetext[1]{Corresponding author: Xianzhong Zhao (x.zhao@tongji.edu.cn)}
\footnotetext[2]{Corresponding author: Ruifeng Luo (ruifengluo.tj@gmail.com)}

\begin{abstract}
Large language models (LLMs), as a novel information technology, are seeing increasing adoption in the Architecture, Engineering, and Construction (AEC) field. They have shown their potential to streamline processes throughout the building lifecycle. However, the robustness and reliability of LLMs in such a specialized and safety-critical domain remain to be evaluated. To address this challenge, this paper establishes AECBench, a comprehensive benchmark designed to quantify the strengths and limitations of current LLMs in the AEC domain. The benchmark features a five-level, cognition-oriented evaluation framework (i.e., Knowledge Memorization, Knowledge Understanding, Knowledge Reasoning, Knowledge Calculation, and Knowledge Application). Based on the framework, 23 representative evaluation tasks were defined. These tasks were derived from authentic AEC practice, with scope ranging from codes retrieval to specialized documents generation. Subsequently, a 4,800-question dataset encompassing diverse formats, including open-ended questions, was crafted primarily by engineers and validated through a two-round expert review. Furthermore, an “LLM-as-a-Judge” approach was introduced to provide a scalable and consistent methodology for evaluating complex, long-form responses leveraging expert-derived rubrics. Through the evaluation of nine LLMs, a clear performance decline across five cognitive levels was revealed.  Despite demonstrating proficiency in foundational tasks at the Knowledge Memorization and Understanding levels, the models showed significant performance deficits, particularly in interpreting knowledge from tables in building codes, executing complex reasoning and calculation, and generating domain-specific documents. Consequently, this study lays the groundwork for future research and development aimed at the robust and reliable integration of LLMs into safety-critical engineering practices. The code and dataset are available at \url{https://github.com/ArchiAI-LAB/AECBench}.

\end{abstract}


\section{Introduction}
\label{sec1}
\begin{figure}[!tbp]
	\centering
	\includegraphics[width=\textwidth]{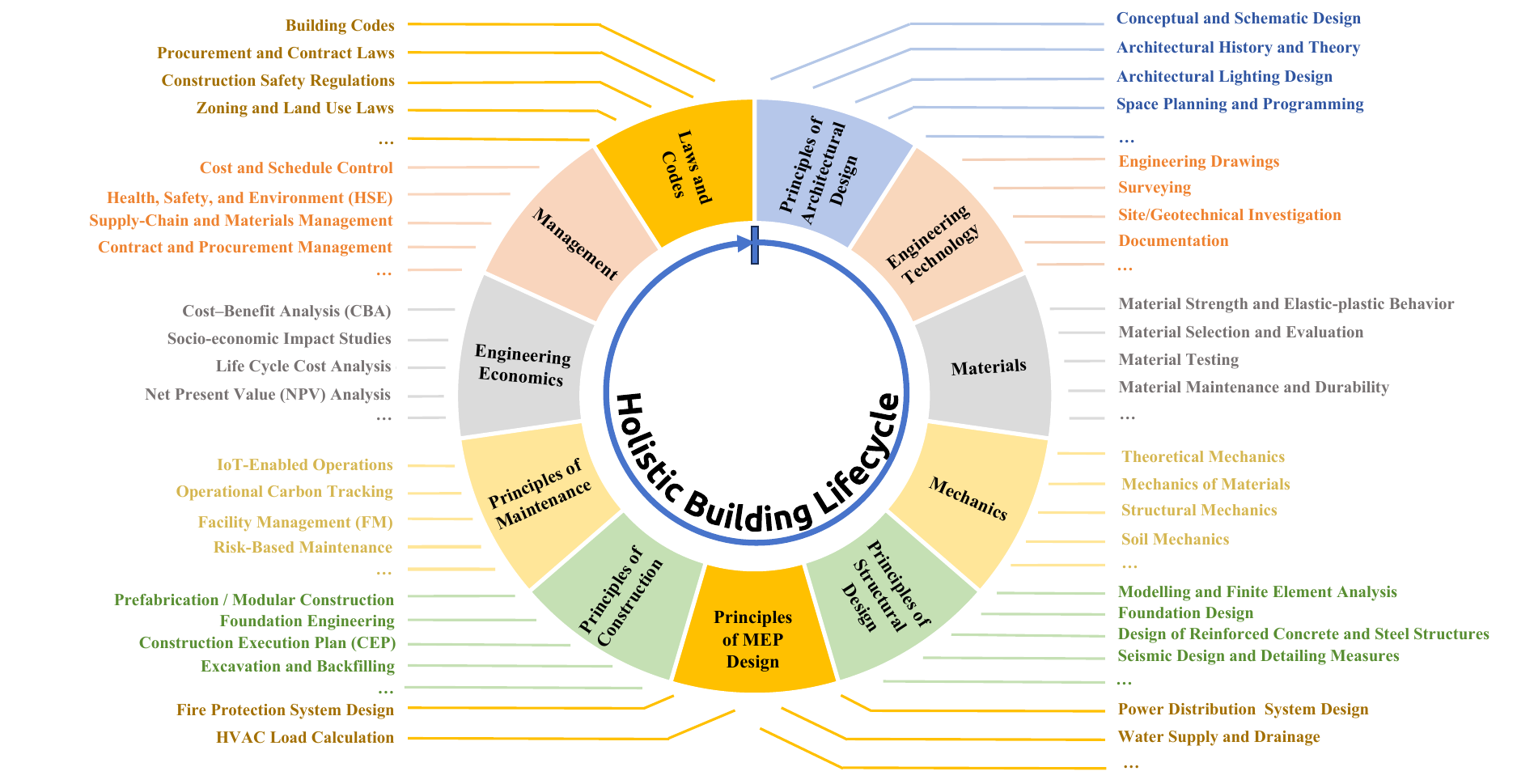}
	\caption{Building lifecycle knowledge dimensions}
	\label{fig:knowledge cycle}
\end{figure}

The Architecture, Engineering, and Construction (AEC) industry is characterized by its inherent complexity and highly specialized, safety-critical nature. As an interdisciplinary convergence of civil engineering and allied fields, the domain demands extensive knowledge and operational capabilities spanning the entire project lifecycle~\cite{zhao_ai_2022}. Figure~\ref{fig:knowledge cycle} illustrates the multifaceted knowledge required in this integrative paradigm, encompassing 11 key domains, as well as numerous specialized topics radiating from these central domains. These domains are highly interrelated and collectively contribute to successful project completion. For instance, structural design is not solely governed by the principles of mechanics; it must also integrate considerations of architectural aesthetics, material properties, construction techniques, fire safety, energy efficiency, and environmental regulations. This deep interconnectivity places stringent demands on the accuracy and reliability of engineering practices.

While large language models (LLMs) offer the potential to augment these knowledge-intensive tasks by leveraging vast cross-domain repositories~\cite{rane_chatgpt_2023,ghimire_opportunities_2024,kampelopoulos_review_2025}, deploying them in such high-stakes environments necessitates rigorous validation. Therefore, establishing a standardized and domain-specific evaluation benchmark is essential to ensure their reliability and proficiency in these contexts~\cite{10.1145/3641289,ITU-T_F.748.44_2025,huang2023cevalmultilevelmultidisciplinechinese,63f50ad476b04d50bb1a6c7459419e96}. However, a comprehensive evaluation framework that fully aligns with the intricate cognitive and operational demands of the AEC industry is currently absent.

To address this critical gap, this paper presents AECBench, a comprehensive Chinese benchmark specifically designed to evaluate LLMs in real-world AEC scenarios across diverse cognitive levels. The main innovations are as follows:

\begin{itemize}
\item \textbf{Hierarchical Cognitive Framework}: A novel hierarchical cognitive framework tailored for the AEC domain is introduced. This framework is designed to mirror the graduated level of cognitive demands found in AEC tasks, from basic knowledge memorization to complex application. The framework is structured into five distinct levels. Notably, it includes dedicated levels for \textit{Knowledge Reasoning} and \textit{Knowledge Calculation}, which are two critical skills in the AEC domain. For a more fine-grained assessment of practical abilities, the \textit{Knowledge Application} level is further partitioned into three sub-levels: \textit{Analysis}, \textit{Evaluation}, and \textit{Creation}. A more comprehensive and realistic evaluation of an LLM’s capabilities is made possible by this hierarchical design, through which its strengths and weaknesses across different cognitive levels can be identified.

\item \textbf{High-Quality Benchmark Dataset}: A high-quality benchmark dataset consisting of 4,800 meticulously crafted or curated questions across 23 evaluation tasks was constructed, characterized by three core strengths. First, the evaluation tasks were curated by domain engineers to reflect real-world scenarios in the AEC field, with questions independently crafted or curated from building codes, proprietary internal documents, and licensure examinations. Second, the dataset offers a diverse range of question formats, including multiple-choice, classification, extraction, and open-ended generation, designed to authentically reflect the complexity and dynamic, open-ended nature of real-world AEC scenarios and professional workflows. Third, each question was subjected to multiple rounds of data review by top professionals from leading AEC enterprises, ensuring high clarity and accuracy throughout the dataset.

\item \textbf{Automated Evaluation Pipeline and Open-source Release}: An automated evaluation pipeline for open-ended, long-form responses is introduced. It employs an "LLM-as-a-judge" approach to score documents against the rubrics established by domain experts. This method significantly enhances evaluation scalability and speed while ensuring that the assessment remains grounded in expert knowledge. The entire AECBench, including the code and dataset, is to be released as an open-source resource for the community.
\end{itemize}

\section{Related works}
\subsection{LLM integration in AEC lifecycle}
Recent research demonstrates the extensive integration of LLMs across the AEC lifecycle, revealing a shift from simple queries to complex, agentic workflows. In the design phase, applications range from automating the generation of building energy models (BEMs)~\cite{jiang_eplus-llm_2024} to enabling conversational interactions with Building Information Models (BIMs)~\cite{zheng_dynamic_2023} and converting textual codes into executable rules for automated compliance checking~\cite{yang_prompt-based_2024, chen_automated_2024}. In structural engineering, LLMs function as controllers for generative design~\cite{RN23} and analysts for structural health monitoring data~\cite{RN22}. During construction and operation, applications extend to project scheduling automation~\cite{prieto_investigating_2023} and safety management, including accident report analysis~\cite{smetana_highway_2024} and retrieval-augmented safety assistance~\cite{uhm_effectiveness_2025}. However, these studies currently rely on ad-hoc evaluations within specific, narrow scenarios. Consequently, there lacks a standardized benchmark to rigorously validate whether a model possesses the generalized and foundational capabilities required to support these diverse, high-stakes workflows.

\subsection{Development of LLM evaluation benchmarks}
Domain-specific benchmarks serve as critical infrastructure for the LLM ecosystem, offering quantitative metrics that not only validate model robustness but also guide model selection, technical optimization, and regulatory governance~\cite{10.1145/3641289,ITU-T_F.748.44_2025, huang2023cevalmultilevelmultidisciplinechinese}. Reflecting this significance, dedicated evaluation frameworks have been established in high-stakes domains such as law, medicine, and finance~\cite{fei2023lawbenchbenchmarkinglegalknowledge,zhu2023promptcbluechineseprompttuning,guo2024finevalchinesefinancialdomain}. However, the AEC industry is characterized by its unique multidisciplinary convergence, requiring not only legal-analogous code comprehension but also financial-analogous quantitative proficiency and domain-specific analysis. Consequently, the direct adaptation of frameworks from other domains is inherently insufficient.

In the specific context of the AEC domain, several initial attempts have been made to bridge this gap. Qin et al. curated a benchmark comprising 100 questions derived from textbook content~\cite{Qin2023LLMConstruction}, while Wu et al. created a larger dataset of 875 questions covering architectural design and planning~\cite{Wu_Jiang_Fan_Li_Xu_Zhao_2025}. Despite offering valuable initial contributions, these benchmarks suffer from fragmented coverage confined to narrow sub-domains or single topics.

To achieve more comprehensive coverage, subsequent benchmark research has gravitated towards professional licensure examinations spanning multiple disciplines within the AEC domain. Liang et al. developed aice.AEC-Bench, utilizing over 1,000 questions from licensure examinations to assess abilities across five cognitive levels~\cite{chen_liang_aiceaec-bench_2025}. Their findings reveal a progressive performance decline from lower-order cognitive tasks to higher-order ones. Similarly, Zhang et al. curated Civil-Eval, a collection of 426 single-choice and 91 multiple-choice items spanning eight subjects. These questions are rigorously classified into 'simple' and 'hard' groups~\cite{civileval}. Continuing this line of research, Zhou et al. established Civil-Bench, a larger benchmark covering 13 categories of professional engineering exam questions, including 14,823 objective questions and 269 subjective questions~\cite{civilbench}. However, a key limitation persists across these benchmarks. While Civil-Bench incorporates subjective formats, it remains overwhelmingly dominated by objective problems, and aice.AEC-Bench and Civil-Eval exclusively rely on choice-based questions. Consequently, these examination-based datasets diverge considerably from real-world AEC scenarios, where professionals are frequently required to address open-ended and ill-structured problems.

\section{The design methodology of \bench{}}
\label{sec2}

\subsection{Hierarchical evaluation framework}
This section aims to comprehensively and effectively assess the capabilities of large language models by reflecting the unique knowledge, application scenarios, and cognitive challenges inherent to the AEC domain. Challenging the conventional approach of categorizing tasks by their disciplines or difficulties~\cite{huang2023cevalmultilevelmultidisciplinechinese,Qin2023LLMConstruction, Wu_Jiang_Fan_Li_Xu_Zhao_2025}, this paper argues for a more nuanced evaluation framework for the complex AEC field. The rationale is twofold: (i) performance is dictated more by a task's cognitive demands than its subject matter, and (ii) AEC tasks themselves are rarely cognitively monolithic. For example, a task in "Design of Reinforced Concrete and Steel Structures" necessitates memorizing code provisions, understanding load-bearing principles, reasoning about structural integrity, and creating a compliant design document. To adequately capture this complexity, this paper establishes a hierarchical framework grounded in these distinct cognitive abilities, referencing Bloom's Taxonomy. Bloom’s Taxonomy, first introduced in 1956 and revised in 2001, is a well-established cognitive taxonomy~\cite{1956Taxonomy,krathwohl2002revision}. It categorizes learning objectives into six progressive levels: Remember, Understand, Apply, Analyze, Evaluate, and Create. Due to its hierarchical structure, which maps a progression from foundational knowledge to advanced problem-solving, it provides a robust theoretical basis for the proposed evaluation framework. 

\begin{figure}[!tbp]
	\centering
	\includegraphics[width=0.8\textwidth]{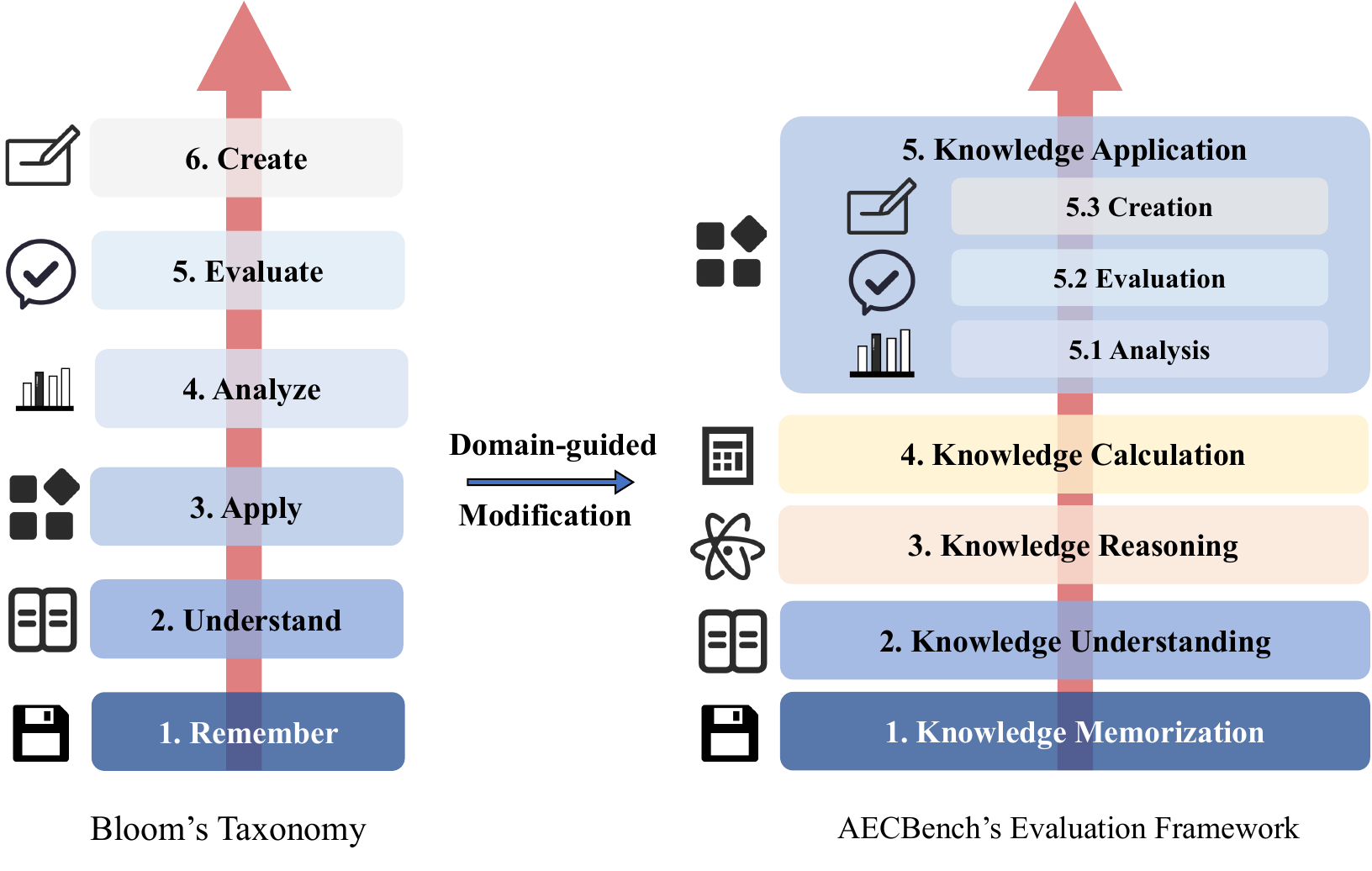}
	\caption{The hierarchical evaluation framework}
	\label{fig:framework}
\end{figure}

The hierarchical evaluation framework consists of five distinct levels: \textit{Knowledge Memorization}, \textit{Knowledge Understanding}, \textit{Knowledge Reasoning}, \textit{Knowledge Calculation}, and \textit{Knowledge Application} (see Figure~\ref{fig:framework}). The proposed framework adapts and extends Bloom’s Taxonomy to cater to the specific demands of the AEC domain. Specifically, it retains the foundational levels of Remember and Understand (i.e., Knowledge Memorization and Knowledge Understanding), while grouping the higher-order skills, Analyze, Evaluate, and Create, as distinct sub-levels within a comprehensive Knowledge Application level. Furthermore, to address the unique requirements of the field, two dedicated levels are introduced: Knowledge Reasoning and Knowledge Calculation. This extension is motivated by the prevalence of tasks requiring logical deduction or numerical computation across key AEC domains (as shown in Figure~\ref{fig:knowledge cycle}), a necessity that is particularly prominent in areas such as mechanics, principles of structural design, and engineering economics.

The Knowledge Memorization level assesses the model's ability to recall foundational AEC facts. The Knowledge Understanding level evaluates LLMs’ capacity to not only retain but also meaningfully interpret and contextualize foundational AEC knowledge, bridging memorized facts with their practical significance and logical connections. The Knowledge Reasoning level assesses an LLM’s ability to utilize its internalized domain knowledge, synthesizing multiple facts, codes, and contextual factors to reach sound, defensible conclusions. This level assesses the model's ability to decompose intricate problems as well as to identify and reconcile relevant constraints. Furthermore, it evaluates whether the model can construct a clear, step-by-step logical chain that leads to viable decisions and judgments. The Knowledge Calculation level assesses an LLM’s ability to apply AEC knowledge to produce verifiable quantitative results. It evaluates whether the LLM can select appropriate formulas, use correct parameters, maintain dimensional consistency, and carry out multi-step calculations. The Knowledge Application level is uniquely tailored to the applied nature of the AEC discipline, and structured into three distinct sub-levels: \textit{Analysis}, \textit{Evaluation}, and \textit{Creation}. Progressing through three ascending levels, it encompasses: Analysis, in which the model inductively extracts, classifies, or verifies domain-specific data or information; Evaluation, where it assesses whether a given document or solution meets the AEC sector’s requirements; and Creation, where it synthesizes and produces domain-specific documents such as design proposals, technical specifications, or construction schedules that are both novel and technically sound. 

Notably, the Knowledge Reasoning and Knowledge Calculation levels assess abilities that share conceptual similarities with those evaluated in other major benchmarks. For instance, these abilities are similar to the mathematical reasoning skills tested in prominent benchmarks like GSM8K~\cite{RN9} and Math-500~\cite{lightman2023letsverifystepstep}. These conceptual similarities are also evident when compared to domain-specific benchmarks such as FinanceMath~\cite{zhao-etal-2024-knowledgefmath}, MedCalc-Bench~\cite{khandekar2024medcalcbench}, and LawBench~\cite{fei2023lawbenchbenchmarkinglegalknowledge} for the financial, medical, and legal domains, respectively. This conceptual alignment with established benchmarks underscores the framework's focus on critically recognized LLM competencies.

\subsection{Task design}
\begin{figure}[!tbp]
	\centering
	\includegraphics[width=\textwidth]{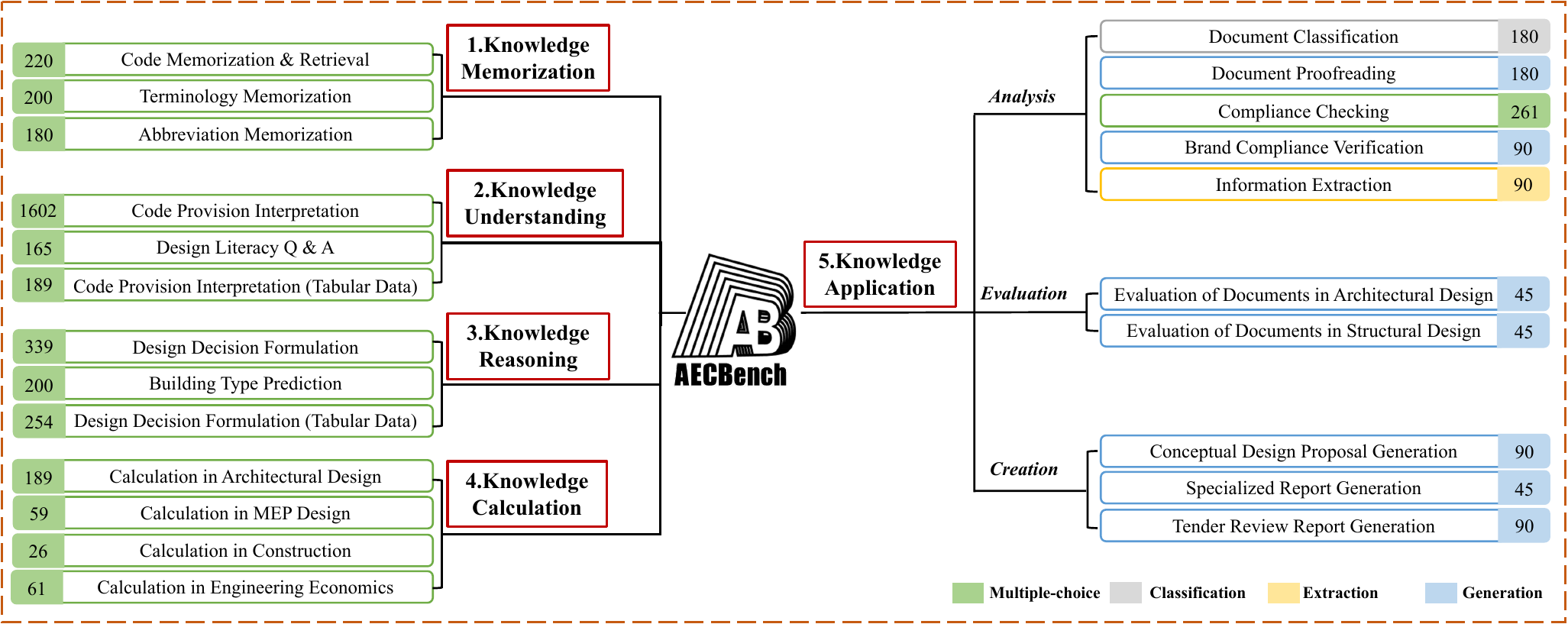}
	\caption{Designed evaluation tasks of AECBench}
	\label{fig:tasks}
\end{figure}

To ensure the comprehensive representativeness of the benchmark, the 23 tasks were selected to reflect both the breadth of the AEC project lifecycle and the depth of cognitive skills required in the field. The tasks capture breadth by systematically covering core professional activities from conceptual design and regulatory management to engineering and construction. Simultaneously, they capture depth by mapping these activities to the five cognitive levels (see Figure~\ref{fig:tasks})—ranging from foundational knowledge to complex application. These tasks were all derived from real-world AEC scenarios and were curated by domain engineers to ensure practical relevance.

\subsubsection{Knowledge Memorization tasks}

This paper establishes three tasks designed to assess the memorization of key AEC knowledge, specifically targeting three typical types: (i) core law articles and code provisions, (ii) fundamental terminology, and (iii) domain-specific abbreviations. These three types were selected for their universal representativeness: Core law articles and code provisions constitute the industry's mandatory regulatory and safety bedrock. Fundamental terminology provides the essential vocabulary for all unambiguous technical communication between architects, engineers, and contractors. Domain-specific abbreviations are the high-frequency, practical shorthand used universally across all project documentation, from technical drawings to reports.
\begin{itemize}
  \item \textbf{Code Memorization \& Retrieval} (1-1): Given an exact excerpt from a provision of building codes, identify the originating code by selecting the correct option from four candidates.
  \item \textbf{Terminology Memorization} (1-2): Given a definition of a domain-specific term, identify the correct terminology corresponding to the definition by selecting the correct option from four candidates.
  \item \textbf{Abbreviation Memorization} (1-3): Given a domain-specific abbreviation, identify its correct expansion by selecting the correct option from four candidates.
\end{itemize}

\subsubsection{Knowledge Understanding tasks}

\begin{figure}[!tbp]
	\centering
	\includegraphics[width=\textwidth]{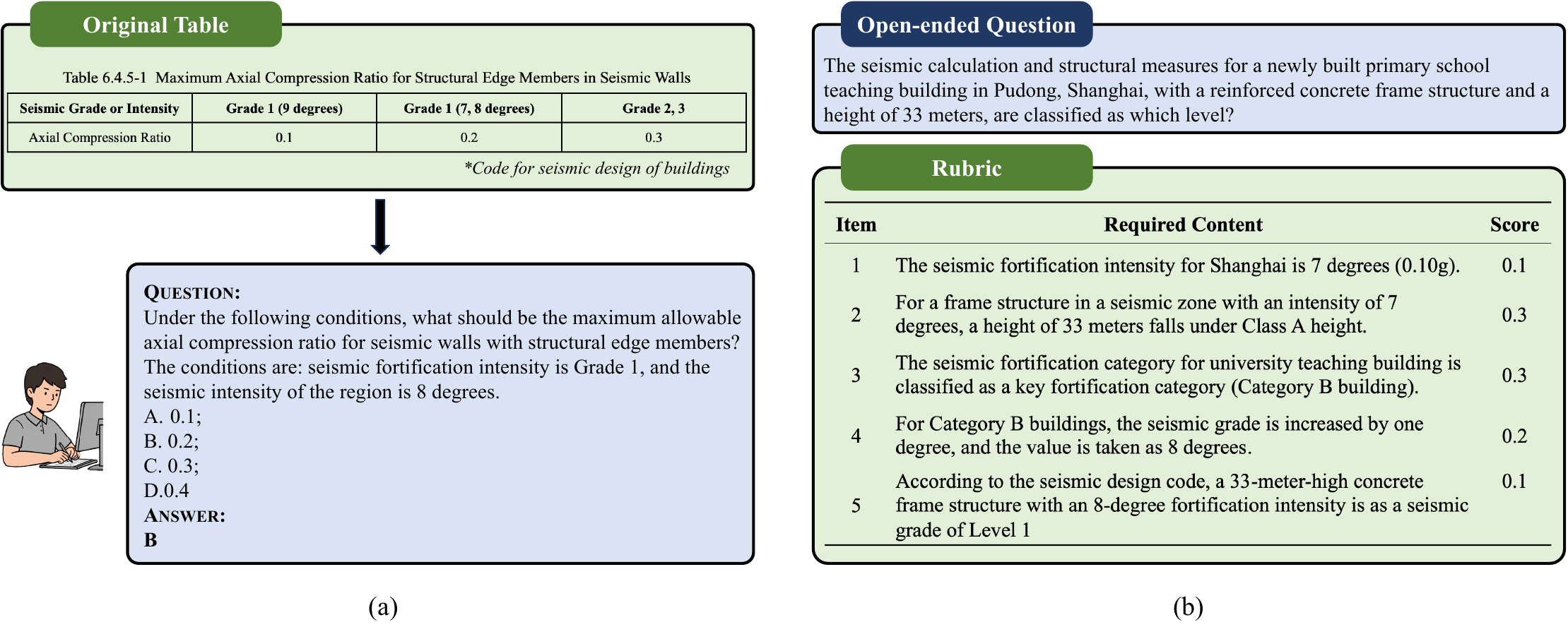}
    \captionsetup{justification=centering}
	\caption{Illustrative examples of the two proposed tasks
    \\
           \textit{(a) the code provision interpretation (tabular data) task; (b) the design proposal generation task}}
	\label{fig:table data}
\end{figure}

To assess the comprehension of knowledge from two critical sources, this paper establishes two types of tasks: one focused on standards and codes, and the other on established design literacy. The comprehension of standards and codes tests the core professional skill of interpreting specific requirements, both from complex text and from the dense tabular data ubiquitous in engineering documents. Established design literacy represents the complementary, fundamental body of engineering principles and concepts required for practical application.
\begin{itemize}
  \item \textbf{Code Provision Interpretation} (2-1): Given a partially paraphrased description related to a code provision, the model must demonstrate comprehension of the provision’s qualitative or quantitative requirements by selecting the correct option from four candidates.
  \item \textbf{Design Literacy Q\&A} (2-2): Given an incomplete exposition of domain design literacy, the model must infer the missing core concept and select the correct option from four candidates.  
  \item \textbf{Code Provision Interpretation(Tabular data)}(2-3): Given a partial description related to a table in the code provision, the model must demonstrate comprehension of the provision’s qualitative or quantitative requirements by selecting the correct option from four candidates.
\end{itemize}
It is noted that the code provision interpretation task is further divided into two sub-tasks, and a task (Task 2-3) that requires the model to comprehend knowledge contained within code tables is designed. Figure~\ref{fig:table data}(a) illustrates an example of the question generation process from tabular data. In the figure, a multiple-choice question was crafted based on Table 6.4.5-1 of "Code for seismic design of buildings"~\cite{GB50011-2010}, requiring the model to select the maximum axial compression ratio for a seismic wall under specific conditions.

\subsubsection{Knowledge Reasoning tasks}
To evaluate LLMs' abilities of reasoning with knowledge to make sound, compliant judgments, this level includes two representative tasks designed to evaluate LLMs' logical reasoning abilities: a deductive task that requires making design decisions based on established knowledge, and an inductive task that involves inferring a building's type from its description. Deductive reasoning (Design Decision Formulation) is the critical process of applying complex rules and standards to specific project scenarios. This skill is paramount as it directly impacts engineering safety and code compliance. Concurrently, inductive reasoning (Building Type Inference) assesses the vital diagnostic skill of making preliminary judgments about a project. This capacity is a fundamental prerequisite for countless downstream tasks, as professionals must first correctly identify a project's type (e.g., residential, hospital, industrial) before they can apply the correct set of standards or make sound design decisions.
\begin{itemize}
  \item \textbf{Design Decision Formulation} (3-1): Given a descriptive scenario from a project (e.g., project specifications, site conditions), employ deductive reasoning to determine which of four candidates, representing possible measures, action decisions, or patterns of change, is both logically coherent and compliant with the applicable codes and standards.
  \item \textbf{Building Type Inference} (3-2): Given a descriptive passage about a specific building, apply inductive reasoning to infer its building type and select the correct option from four candidates.
  \item \textbf{Design Decision Formulation (Tabular Data)} (3-3): This task is identical to Task 3-1, except that it requires referencing knowledge contained in the code tables.
\end{itemize}
Similar to the task design process in the Knowledge Understanding level, the design decision formulation task is further divided into two sub-tasks based on whether tabular code knowledge is required.
\subsubsection{Knowledge Calculation tasks}
To assess the quantitative ability to execute precise, verifiable calculations, the four selected tasks were designed to be comprehensively representative, capturing how this essential skill manifests ubiquitously across the entire building lifecycle. This includes the core calculations required in architectural and MEP design, the practical mathematics of the construction phase, and the fundamental financial analysis of engineering economics. These tasks collectively assess the critical numerical literacy required in the industry.
\begin{itemize}
  \item \textbf{Calculation in Architectural Design} (4-1): Given a quantitatively formulated problem in the domain of architectural and structural design (e.g., calculating the moment of inertia of a cross-section), perform the requisite numerical analyses to determine the single correct solution from four candidates.
  \item \textbf{Calculation in MEP Design} (4-2): This task is identical to Task 4-1, but it focuses on assessing calculation abilities in MEP design (e.g., calculating the airflow rate of a fan).
  \item \textbf{Calculation in Construction} (4-3): This task is identical to Task 4-1, but it focuses on assessing calculation abilities in construction (e.g., calculating the project duration and worker shift scheduling).
  \item \textbf{Calculation in Engineering Economics} (4-4): This task is identical to Task 4-1, but it focuses on assessing calculation abilities in engineering economics (e.g., calculating the interest and net present value).
\end{itemize}

\subsubsection{Knowledge Application tasks}
At the Knowledge Application level, multiple representative tasks were designed for each sub-level. Five representative application tasks are considered at the Analysis sub-level: Document Classification, Document Proofreading, Compliance Checking, Brand Compliance Verification, and Information Extraction. These tasks assess the essential, high-frequency skills that define daily practice. This includes not just managing information flow by classifying and categorizing documents or extracting entities from tenders, but the critical, high-stakes processes of performing calculations and reasoning to check code compliance and detecting brand-related deviations in bid submissions. At the Evaluation sub-level, the tasks assess the fundamental competency of acting as a rubric-based judge to score professional documents, a vital simulation of the industry's critical peer-review and quality assurance loops. At the Creation sub-level, this paper designs tasks that require LLMs to generate high-value, structured documents that propel a project, such as the conceptual design proposals that initiate the planning phase, the specialized reports that define the detailed design, and the tender review reports from the procurement phase that summarize discrepancies.

\paragraph{Analysis}

\begin{itemize}

  \item \textbf{Document Classification} (5-1-1): Given a sentence extracted from a domain-specific document, determine and assign the sentence to its appropriate professional discipline category.
  \item \textbf{Document Proofreading} (5-1-2): Given a sentence drawn from a domain-specific document, identify and rectify all linguistic defects, including spelling mistakes, redundancies, omissions, word-order errors, and semantic inaccuracies, and produce a coherent, grammatically correct, and semantically faithful revision.
  \item \textbf{Compliance Checking} (5-1-3): Given a design case with multiple parameters, the model is required to analyze whether the design complies with code provisions through calculations and logical reasoning. If non-compliant, the model must identify the specific cause of the violation. Finally, select the correct violation reason from four candidates.
  \item \textbf{Brand Compliance Verification} (5-1-4): Given a tabular equipment specification sheet, detect and report all potential brand-related deviations present in the corresponding bid submissions.
  \item \textbf{Information Extraction} (5-1-5): Given a long, domain-specific document (e.g., a tender announcement), identify and extract every entity mention that matches a predefined inventory of entity types.
 
\end{itemize}
\paragraph{Evaluation}
\begin{itemize}
  \item \textbf{Evaluation of Documents in Architectural Design} (5-2-1): Given a design scenario and an architectural proposal, the model is required to act as a judge and score the proposal based on predefined criteria, including content accuracy, professional expression, readability, and formatting.
  \item \textbf{Evaluation of Documents in Structural Design} (5-2-2): This task is identical to Task 5-2-1, but it focuses on evaluating documents in structural design.
\end{itemize}
\paragraph{Creation}
\begin{itemize}
  \item \textbf{Conceptual Design Proposal Generation} (5-3-1): Given a project brief and its associated design specifications, automatically generate a preliminary design proposal.
  \item \textbf{Specialized Report Generation} (5-3-2): Given the essential project brief, automatically generate a structural design basis report that adheres to established codes.
  \item \textbf{Tender Review Report Generation} (5-3-3): Given a device specification file and an equipment specification table, the model must first identify all parameter mismatches and then generate a coherent tender review report that summarizes these discrepancies.
\end{itemize}

\subsection{Question design}
\subsubsection{Diverse question formats in AECBench}

AECBench employs a diverse set of question formats including multiple-choice, classification, extraction, and generation. Figure~\ref{fig:tasks} presents the distribution of questions across individual tasks, which are hierarchically organized by cognitive level and color-coded to denote these distinct formats. Multiple-choice questions (MCQs) were used across the Knowledge Memorization, Understanding, Reasoning, and Calculation levels, as well as for the compliance checking task within the Knowledge Application level. They offer clear evaluation metrics (i.e., accuracy) and serve as a simple but effective proxy for assessing advanced capabilities of LLMs. For all tasks within the Knowledge Application level, excluding the compliance checking task, the question formats were intentionally diversified and specifically tailored to the requirements of each task. The classification format was applied to the document classification task; the extraction format was applied to the information extraction task; and a generative format was applied to both the document proofreading and brand compliance verification tasks. It must be mentioned that open-ended questions were utilized for the tasks at the Creation sub-level. Details associated with the open-ended questions are presented in Section~\ref{open-ended}.

By moving beyond a sole reliance on the multiple-choice format, the benchmark mitigates inherent limitations such as option-setting bias and the possibility of correct guessing~\cite{xu2023scsafetymultiroundopenendedquestion}. This approach also enables a more robust evaluation of a model's abilities in authentic AEC practice, which are difficult to assess with multiple-choice questions~\cite{myrzakhan2024openllmleaderboardmultichoiceopenstylequestions,arora2025healthbenchevaluatinglargelanguage}. 

\subsubsection{Open-ended questions and automated rubric-based evaluation}
\label{open-ended}
The proposed AECBench highlights the use of open-ended questions to rigorously assess a model's ability to produce complex and lengthy engineering documents. Instead of simply generating a "yes/no" or predetermined response, the model is required to generate professional documents (for instance, conceptual design proposals) with approximately 2,000 words following an instruction and a prompt. Open-ended questions provide deeper insight into the model's generative capabilities, nuanced contextual understanding, and ability to tackle complex tasks. 

A rubric-based assessment pipeline is established to facilitate the rapid and scalable evaluation of open-ended questions, where LLMs act as automated evaluation agents. Compared to the reference-free evaluation, this rubric-based pipeline anchors the LLM's judgment to objective, expert-defined criteria. This methodology has been suggested to ensure the evaluation is a trustworthy indicator of domain expert opinion, enabling the model-based evaluator to achieve a level of reliability comparable to inter-annotator agreement among human experts~\cite{arora2025healthbenchevaluatinglargelanguage,zheng2023judgingllmasajudgemtbenchchatbot}. This pipeline follows a two-stage workflow. Firstly, the domain experts predefine a detailed set of rubric criteria that serve as the evaluation reference. Subsequently, using the LLM-as-a-Judge approach~\cite{zheng2023judgingllmasajudgemtbenchchatbot}, a generated document is assessed by scoring it against these expert-defined rubrics. 

Figure~\ref{fig:table data}(b) presents an example of the question and its rubric in the specialized report generation task. The example in the figure requires determination of the seismic design parameters for a specific building. The rubric breaks down the problem into five sequential steps, each of which is awarded partial credit, to assess the model's ability to apply codes. This design shifts expert effort from exhaustive grading to the upfront definition of rubrics, thereby leveraging crucial expert knowledge, lowering human workload, and greatly enhancing both scalability and objectivity. 

\begin{figure}[!tbp]
	\centering
	\includegraphics[width=\textwidth]{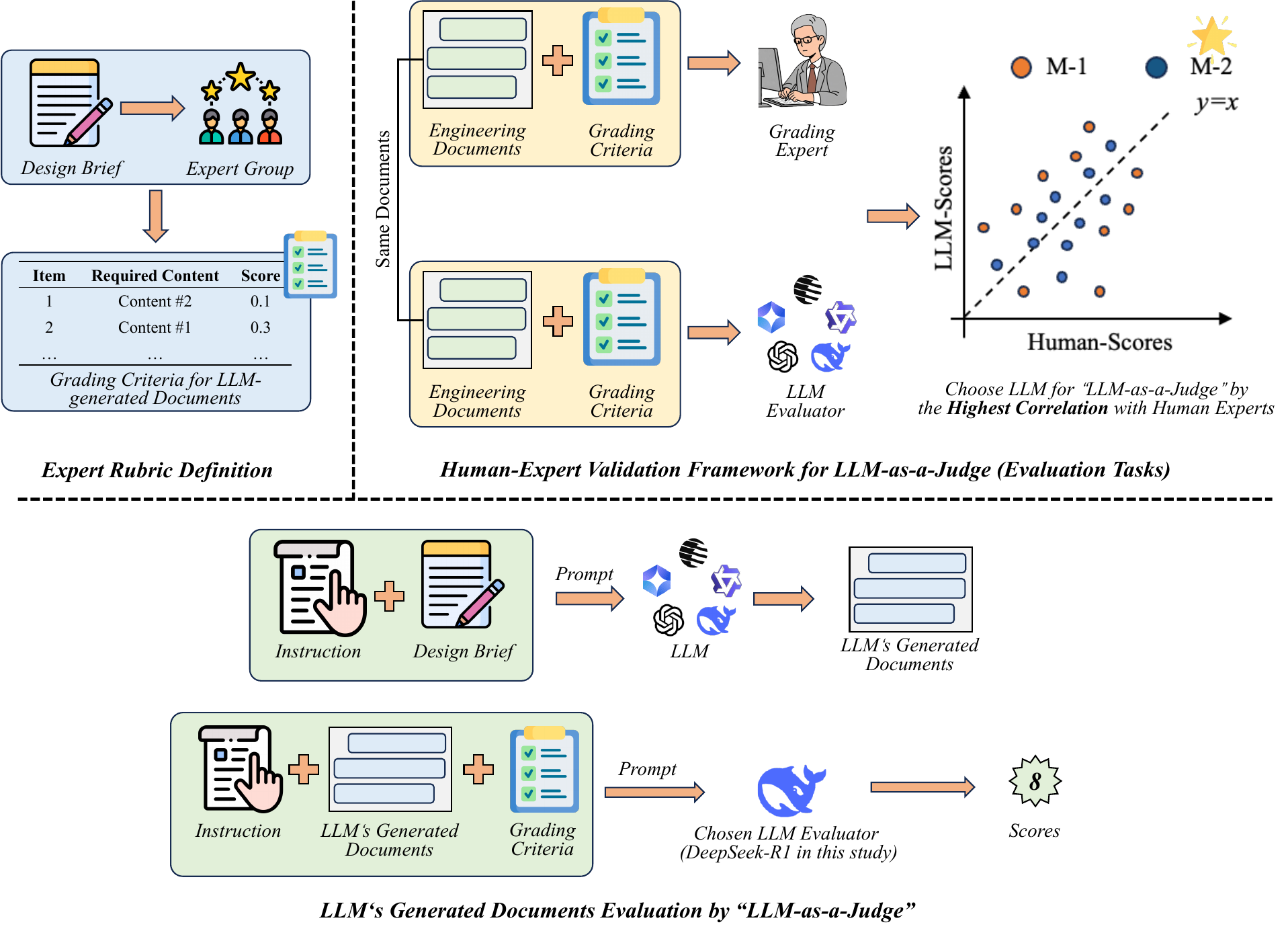}
	\caption{Automated evaluation pipeline for open-ended questions}
	\label{fig:evaluation and creation tasks}
\end{figure}

To validate the reliability of the automated evaluation pipeline, this paper establishes parallel validation tasks in the Evaluation sub-level (Tasks 5-2-1 and 5-2-2). This involves tasks where both human experts and LLMs act as judges to score the same documents. In the Evaluation tasks, the model exhibiting the highest correlation with the scores of human experts is selected and subsequently employed as the judge for the generation tasks. A schematic overview of the assessment framework for Evaluation and Creation tasks is presented in Figure~\ref{fig:evaluation and creation tasks}.

\section{Dataset construction}
\begin{figure}[!tbp]
	\centering
	\includegraphics[width=0.87\textwidth]{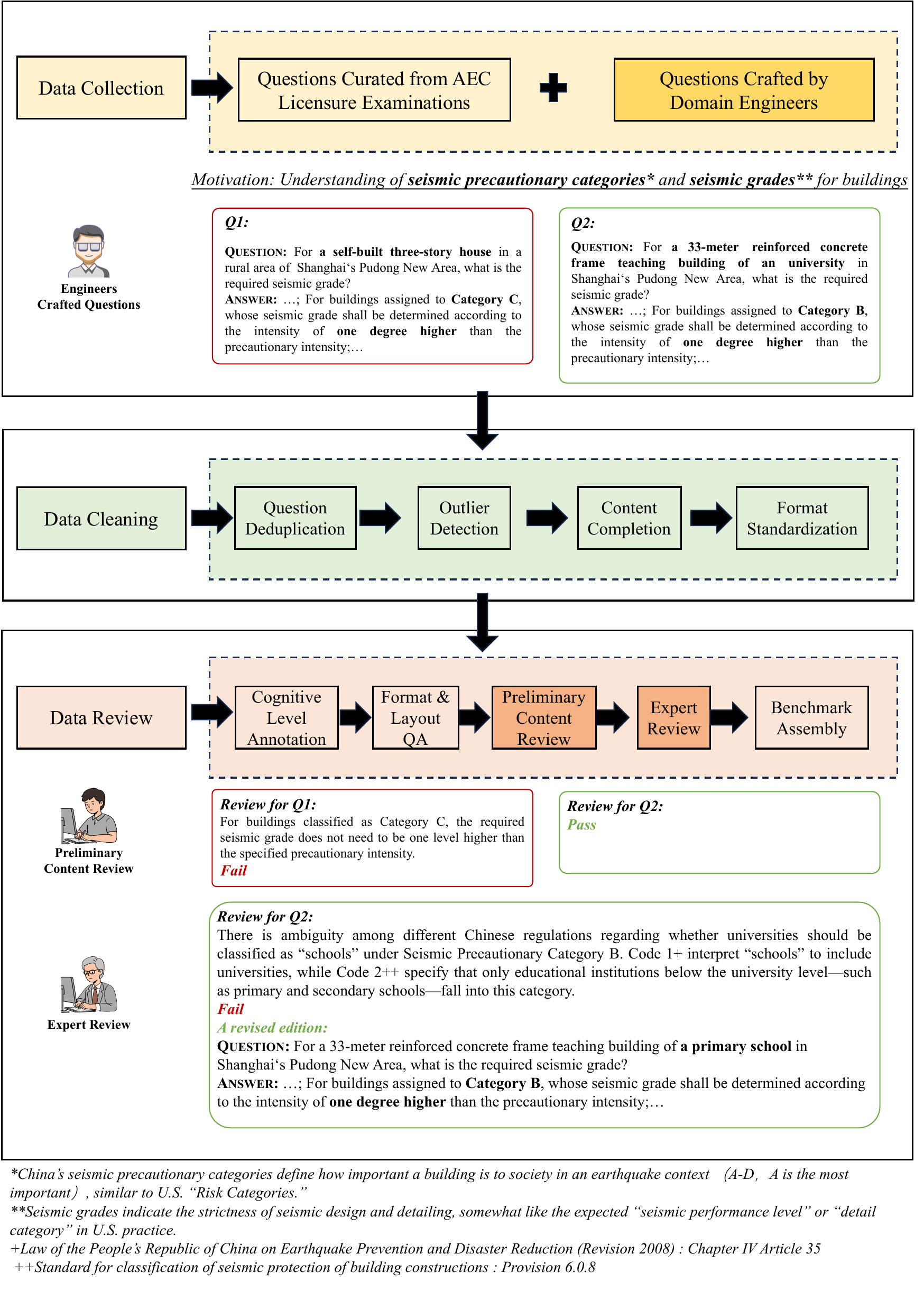}
	\caption{The workflow of constructing the dataset}
	\label{fig:dataset-construction}
\end{figure}

A custom dataset comprising 4,800 questions was constructed based on the evaluation tasks considered in the hierarchical cognitive framework. The dataset was established through three steps: data collection, data cleaning, and data review (see Figure~\ref{fig:dataset-construction}). The data review process highlights a two-round review mechanism (i.e., each individual item is reviewed by a mid-level engineer and then confirmed by an expert), ensuring that a high-quality dataset for LLM evaluation is obtained.

\subsection{Data collection}
The data sources for AECBench evaluation dataset are twofold, with a primary emphasis on questions meticulously crafted by a cross-disciplinary team of domain engineers. Guided by the predefined evaluation framework, the engineers developed targeted assessment items by systematically integrating building codes, proprietary internal materials, and other domain-specific documents, ensuring precise alignment with the cognitive hierarchies and the designed tasks. To ensure foundational coverage, items from official engineering licensure examinations were also included. These questions, sourced from examinations for registered architects, design engineers, and constructors, are valuable for practitioners and address key knowledge across the entire project lifecycle.

\subsection{Data cleaning}
The data cleaning process begins with converting all raw data into a standardized format. This is followed by a multi-step procedure including: question deduplication, which uses n-gram overlap and vector similarity algorithms to remove redundant questions; outlier detection, which focuses on correcting factual and spelling anomalies; and content completion, which verifies the completeness of answers and options. This rigorous process ensures the dataset is reliable and ready for data review.

\subsection{Data review}
The data review highlights a two-round process: preliminary content review and expert review. The preliminary content review is performed by mid-level engineers, who evaluate the reasonableness of each question's content and ensure the overall dataset covers key industry topics. This serves as an effective first line of defense against straightforward inaccuracies. The subsequent stage is the expert review, where senior specialists (engineers with over 10 years of experience) apply their deep expertise to assess the data's representativeness and identify subtle content errors. This expert-led process serves as the core quality control protocol and was found to be critical for identifying errors, clarifying ambiguities, and enhancing overall data reliability. 

Figure~\ref{fig:dataset-construction} visually details the necessity of this two-round data review mechanism through a specific example. In the initial data collection phase, question-and-answer pairs concerning the seismic design for buildings were generated. This review was primarily tasked with checking and correcting foundational, factual errors. As shown with Question Q1, an incorrect statement regarding the seismic requirements for Category C buildings was identified as being in direct contradiction with the building codes. Consequently, Question Q1 was flagged as "Fail". This initial stage served as an effective first line of defense against straightforward inaccuracies. However, a reliance on the preliminary content review alone was determined to be insufficient, as more nuanced issues require deep domain knowledge to be detected. This highlights the decisive importance of the second review stage: expert review. The criticality of this stage is demonstrated by the case of Question Q2. This question was initially assessed as "Pass" during the preliminary content review, as no overt errors were present. However, during the expert review, a critical ambiguity was identified regarding whether a "university" should be classified as a "school" under different interpretations of the code, which affects the required seismic grade. This ambiguity, which could have severely impacted question clarity, was not only identified but also resolved by the domain expert. The question was revised to specify a "primary school" to eliminate any interpretive ambiguity. As a general statement, the two-round iterative mechanism, particularly the inclusion of the second expert-led stage, significantly enhanced the reliability of the final dataset, thereby providing a robust foundation for the ensuing analysis.

\section{Evaluation of large language models}
\label{sec:4}
\subsection{Models}
Nine widely used LLMs were selected, encompassing both open-source and proprietary models, with QwQ-32B, DeepSeek-R1, and GPT o3-mini featuring specialized optimizations for enhanced reasoning capabilities. The selected models and their information are shown in Table~\ref{tab:2}.
\begin{table}[!tbp]
\centering
        \begin{threeparttable} 
        \centering
        \footnotesize
	\caption{Overview of evaluated models}
	\begin{tabular}{cccc}
		\toprule
		Models &Creator & Parameters & Reasoning  \\
		\midrule
            Moonshot-v1-128k \cite{kimiteam2025kimik15scalingreinforcement} & Moonshot AI & Undisclosed &w/o \\
            GLM-4-Plus \cite{glm2024chatglm}	& Zhipu & Undisclosed	&w/o \\
            Qwen-Turbo \cite{qwen2025qwen25technicalreport}& Alibaba &	Undisclosed &w/o \\
            QwQ-32B \cite{qwq32b}& Alibaba &	32B &w/ \\
            DeepSeek-V3 \cite{deepseekai2025deepseekv3technicalreport}	& DeepSeek	& 671B	& w/o \\
            DeepSeek-R1 \cite{deepseekai2025deepseekr1incentivizingreasoningcapability}	& DeepSeek	& 671B	& w/ \\
            GPT-4o \cite{openai2024gpt4ocard}& OpenAI & Undisclosed	& w/o \\
            GPT o3-mini \cite{openai2025competitiveprogramminglargereasoning}& OpenAI & Undisclosed	& w/ \\
            Hunyuan-TurboS \cite{sun2024hunyuanlargeopensourcemoemodel}& Tencent & 560B	& w/o \\
            
		\bottomrule
	\end{tabular}
	\label{tab:2}
         \begin{tablenotes}
           \footnotesize
		\item “w/” (with) or “w/o” (without) reasoning indicates whether the model is able to output its chain-of-thought reasoning or to give only the final answer without exposing intermediate reasoning steps .
        \end{tablenotes} 
        \end{threeparttable}
\end{table}

\subsection{Experiment settings}

\begin{figure}[!tbp]
	\centering
	\includegraphics[width=0.8\textwidth]{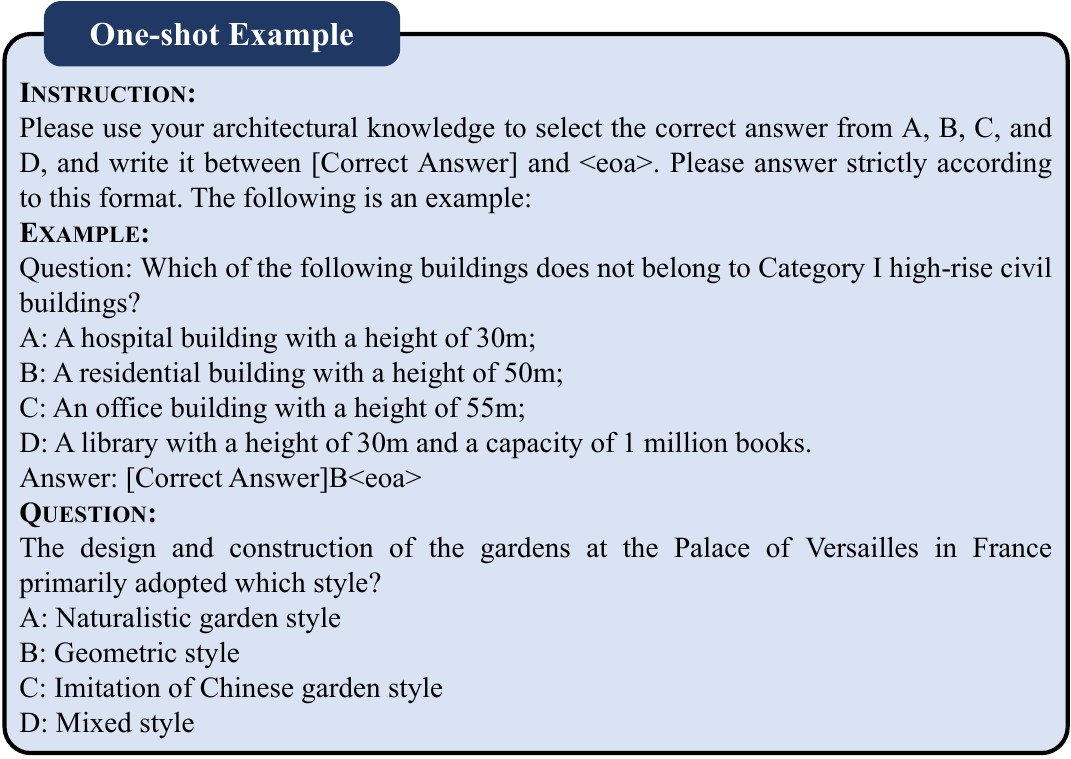}
	\caption{One-shot example for multiple-choice questions}
	\label{fig:5}
\end{figure}

The evaluation experiments were conducted on the OpenCompass platform~\cite{2023opencompass}, which enables automated answer extraction from model outputs and standardized computation of benchmarking metrics through its modular architecture. The experiments were conducted under a one-shot setting, which involves evaluating the model with a single example provided for each task, as shown in Figure~\ref{fig:5}. In this case, the model is expected to understand the task and generalize from this single example with minimal guidance. While instruction-tuned models can perform well in zero-shot conditions~\cite{zhong2023agievalhumancentricbenchmarkevaluating,guo2023evaluatinglargelanguagemodels}, a single example is essential to unambiguously clarify task instructions and output formats across our 23 diverse tasks. More importantly, a few-shot setting was avoided because providing multiple demonstrations has been shown to "befuddle" advanced, instruction-tuned LLMs and, in some cases, even lead to a decline in model performance~\cite{zeng2023measuringmassivemultitaskchinese,li-etal-2024-cmmlu,liu2023m3kemassivemultilevelmultisubject}. Therefore, the one-shot setting provides the optimal balance: it offers the minimal guidance necessary for fair, unambiguous task execution, while ensuring testing the models' internalized domain knowledge rather than their ability to over-fit to in-context examples. All experiments were primarily carried out on a server featuring an NVIDIA 4090 GPU (24GB) and Linux Ubuntu 22.04.3 LTS.

\subsection{Main results}
\subsubsection{General observations}
\begin{figure}[!tbp]
	\centering
	\includegraphics[width=1.0\textwidth]{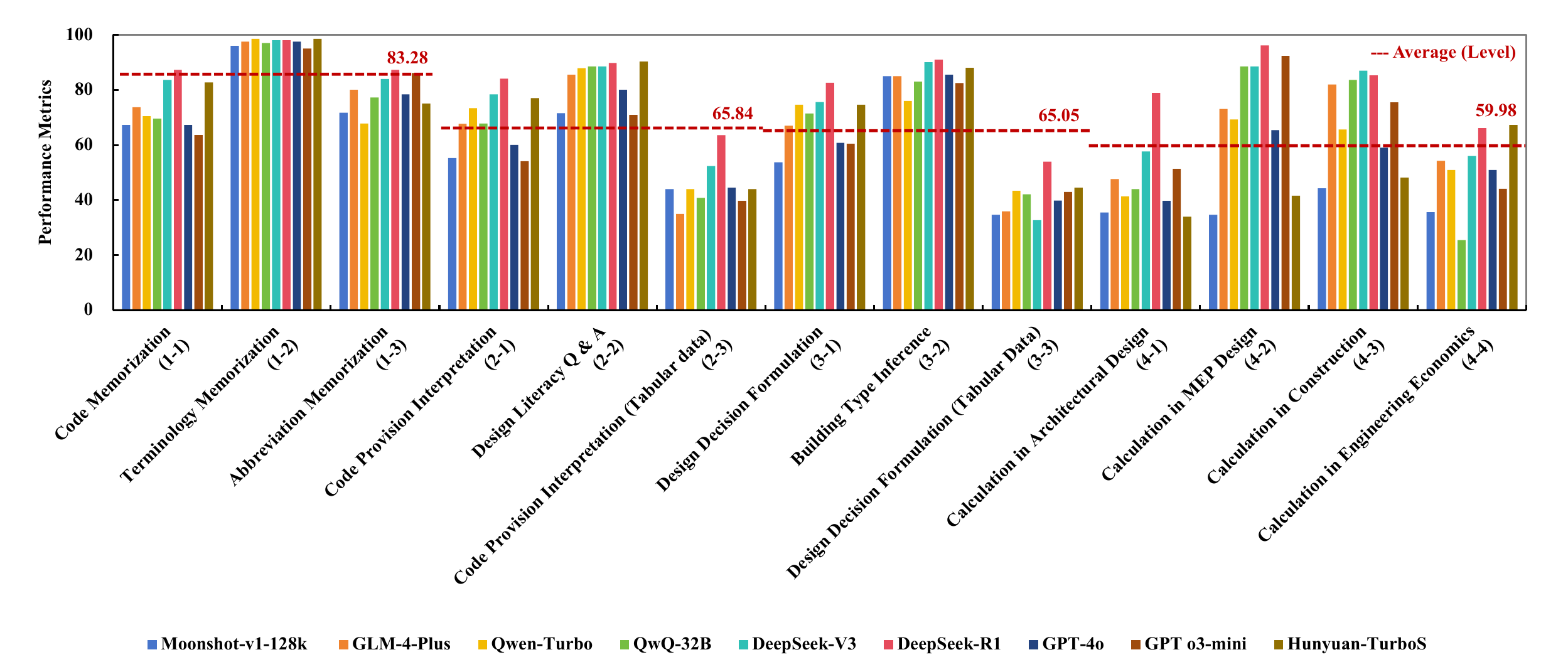}
    \captionsetup{justification=centering}
	\caption{Performance of models in Knowledge Memorization/Understanding/Reasoning/Calculation tasks}
	\label{fig:7}
\end{figure}

\begin{figure}[!tbp]
  \centering
  \includegraphics[width=1.0\textwidth]{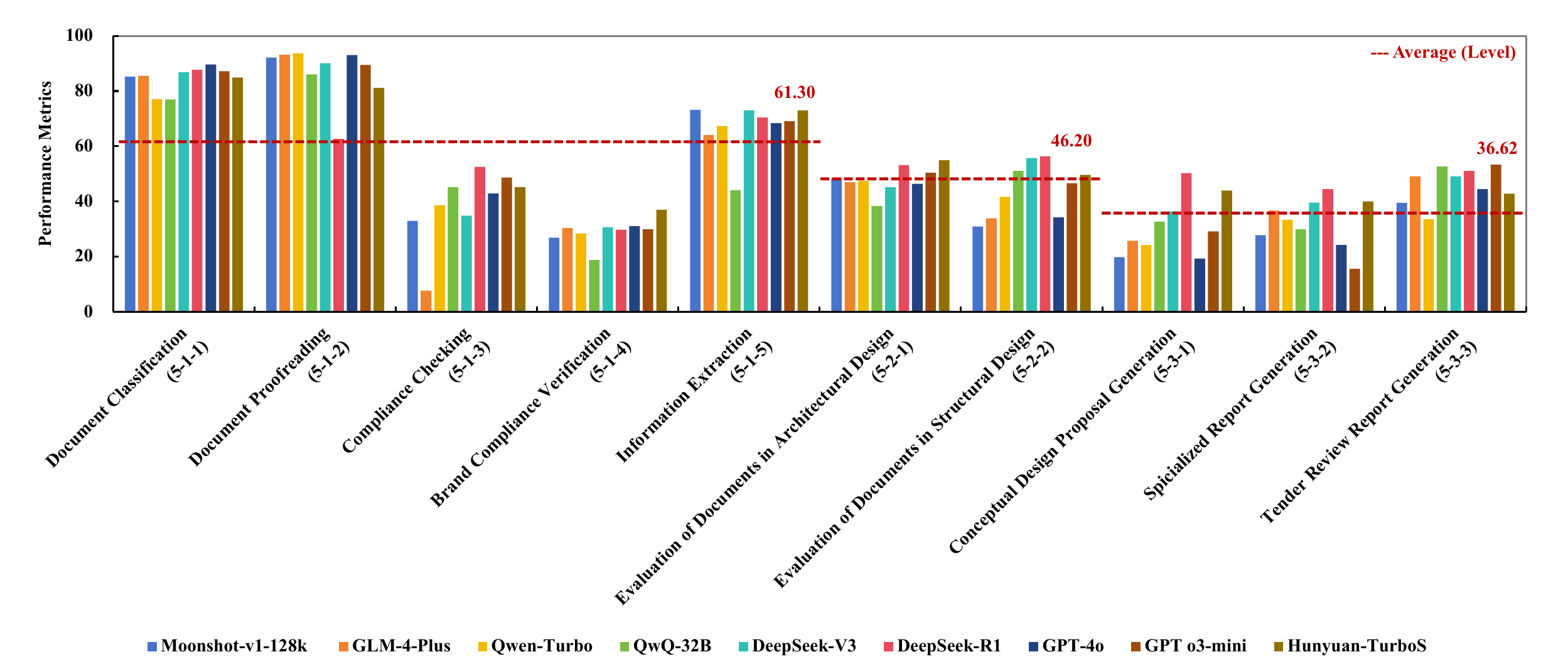}
  \caption{Performance of models in Knowledge Application tasks}
  \label{fig:8}
\end{figure}

A systematic evaluation of nine LLMs was conducted through the evaluation platform. As detailed in Table~\ref{tab:3}, the task-specific performance metrics for each LLM are visualized in Figures~\ref{fig:7} and~\ref{fig:8}. These figures also include horizontal lines representing the average performance for each level and sub-level. The test results clearly suggest a decline in LLMs’ capabilities across the five cognitive levels. A similar downward trend is also evident across the three sub-levels of the Knowledge Application level. Both major inter-level gaps and fine-grained intra-level differences were successfully identified, thereby validating the soundness and necessity of the hierarchical evaluation framework.

\begin{table}[!tbp]
\centering
    \begin{threeparttable} 
    \centering
    \footnotesize                 
\setlength\tabcolsep{4pt}   
\renewcommand{\arraystretch}{1.1}
	\caption{Evaluation results}
    \label{tab:3}
            \begin{tabular}{cccccccccc}
            \toprule
            \textbf{\shortstack{Task\\ID}} & \textbf{\shortstack{Moonshot-v1\\-128k}}& \textbf{\shortstack{GLM-4\\-Plus}}& \textbf{\shortstack{Qwen\\-Turbo}} & \textbf{\shortstack{QwQ\\-32B}} & \textbf{\shortstack{DeepSeek\\-V3}} & \textbf{\shortstack{DeepSeek\\-R1}} & \textbf{\shortstack{GPT\\-4o}} & \textbf{\shortstack{GPT\\o3-mini}}  &\textbf{\shortstack{Hunyuan\\-TurboS}}\\
            \midrule
            1-1 & 67.27 & 73.64 & 70.45 & 69.54 & 83.64 & \textbf{87.27}& 67.27 & 63.64 &82.73 \\
            1-2 & 96& 97.5& \textbf{98.5}& 97& 98&98& 97.5& 95&\textbf{98.5}\\
            1-3 & 71.67 & 80.00 & 67.78 & 77.22 & 83.89 & \textbf{87.22} & 78.33 & 86.11 &75.00 \\
            2-1& 55.24 & 67.67 & 73.41 & 67.73 & 78.34 & \textbf{84.02} & 60.05 & 54.06 &77.03 \\
            2-2& 71.52 & 85.45 & 87.88 & 88.48 & 88.48 & 89.70 & 80.00 & 70.91 &\textbf{90.30}\\
            2-3& 43.92 & 34.92 & 43.92 & 40.74 & 52.38 & \textbf{63.49} & 44.44 & 39.68 &43.92 \\
            3-1& 53.69 & 66.96 & 74.63 & 71.39 & 75.52 & \textbf{82.60} & 60.77 & 60.47 &74.63 \\
            3-2& 85.00 & 85.00 & 76.00 & 83.00 & 90.00 & \textbf{91.00} & 85.50 & 82.50 &88.00 \\
            3-3& 34.65 & 35.83 & 43.31 & 42.13 & 32.68 & \textbf{53.94 }& 39.76 & 42.91 &44.49 \\
            4-1& 35.45 & 47.62 & 41.27 & 43.92 & 57.67 & \textbf{78.84} & 39.68 & 51.32 &33.86 \\
            4-2& 34.62 & 73.08 & 69.23 & 88.46 & 88.46 & \textbf{96.15} & 65.38 & 92.31 &41.53 \\
            4-3& 44.26 & 81.97 & 65.57 & 83.61 & \textbf{86.89} & 85.25 & 59.02 & 75.41 &48.08 \\
            4-4& 35.59 & 54.24 & 50.85 & 25.42 & 55.93 & \textbf{66.10} & 50.85 & 44.07 &67.21 \\
            5-1-1& 85.28 & 85.56 & 77.22 & 76.96 & 86.94 & 87.78 & \textbf{89.72} & 87.22 &85.00 \\
            5-1-2& 92.23 & \textbf{93.24} & 93.75 & 86.12 & 90.19 & 62.73 & 93.14 & 89.51 &81.20 \\
            5-1-3& 32.95 & 7.66 & 38.70 & 45.21 & 34.87 & \textbf{52.49} & 42.91 & 48.66 &45.21 \\
            5-1-4& 26.90 & 30.38 & 28.45 & 18.80 & 30.66 & 29.73 & 31.09 & 29.95 &\textbf{37.00}\\
            5-1-5& \textbf{73.21} & 64.15 & 67.35 & 44.10 & 73.10 & 70.48 & 68.43 & 69.12 &73.10 \\
            5-2-1& 48.4 & 47.1 & 47.4 & 38.3 & 45.2 & 53.2 & 46.5 & 50.5 &\textbf{54.9}\\
            5-2-2& 30.9 & 33.9 & 41.7 & 51.1 & 55.7 & \textbf{56.4} & 34.2 & 46.6 & 49.6 \\
            5-3-1& 19.83 & 25.67 & 24.11 & 32.67 & 36.33 & \textbf{50.28*} & 19.33 & 29.11 &43.94 \\
            5-3-2& 27.78 & 36.67 & 33.33 & 30.00 & 39.56 & \textbf{44.44*} & 24.22 & 15.56 &40.00 \\
            5-3-3& 39.44 & 49.17 & 33.61 & 52.78 & 49.17 & 51.11* & 44.44 & \textbf{53.33}  &42.78 \\
            \bottomrule
            \end{tabular}%

         \begin{tablenotes}
           \footnotesize
		\item The table shows the evaluation results of nine LLMs across twenty-three tasks. All multiple-choice questions are evaluated using accuracy as the metric. For the classification task (Task 5-1-1), the evaluation metric is F1. For Tasks 5-1-4 and 5-1-5, Soft-F1 is used as the evaluation metric, and F0.5 is used as the evaluation metric for Task 5-1-2. For evaluation tasks (Tasks 5-2-1 and 5-2-2), Kendall rank correlation coefficient ($\tau$) is used as the evaluation metric. Kendall's $\tau$ is a statistic used to measure the ordinal association between two measured quantities. It ranges from -1 to +1, where +1 indicates perfect agreement between the two rankings, -1 indicates perfect disagreement, and a value near 0 indicates no correlation. In Tasks 5-3-1, 5-3-2, and 5-3-3, given that the DeepSeek-R1 model serves as the expert model for benchmark scoring, all corresponding performance metrics are annotated with an asterisk (*) to alert readers to potential evaluation bias risks. Though not explicitly shown here, self-enhancement bias~\cite{zheng2023judgingllmasajudgemtbenchchatbot} was analyzed and found to be marginal.To provide a unified and comparable view of performance across all tasks, the results for all metrics have been normalized to a 0-100 scale. As all observed Kendall's $\tau$ values were positive ([0, 1]), they were scaled by 100 (Score = $\tau$ × 100) for clarity and consistent presentation. The numbers in bold in the table represent the best-performing model for each task.
        \end{tablenotes} 
        \end{threeparttable}
\end{table}

A noticeable performance drop was identified between the Knowledge Memorization and Understanding levels, which is attributed to the cognitive leap required to move beyond the retrieval of explicitly encoded facts to the construction of deeper logical relationships. Another steep decline in performance was evident among the sub-levels of knowledge application, particularly in the creation sub-level. This decline is compounded by a methodological factor, as a transition is made from a uniform multiple-choice format to diverse task types with varied evaluation metrics. 

\begin{figure}[!tbp]
	\centering
	\includegraphics[width=\textwidth]{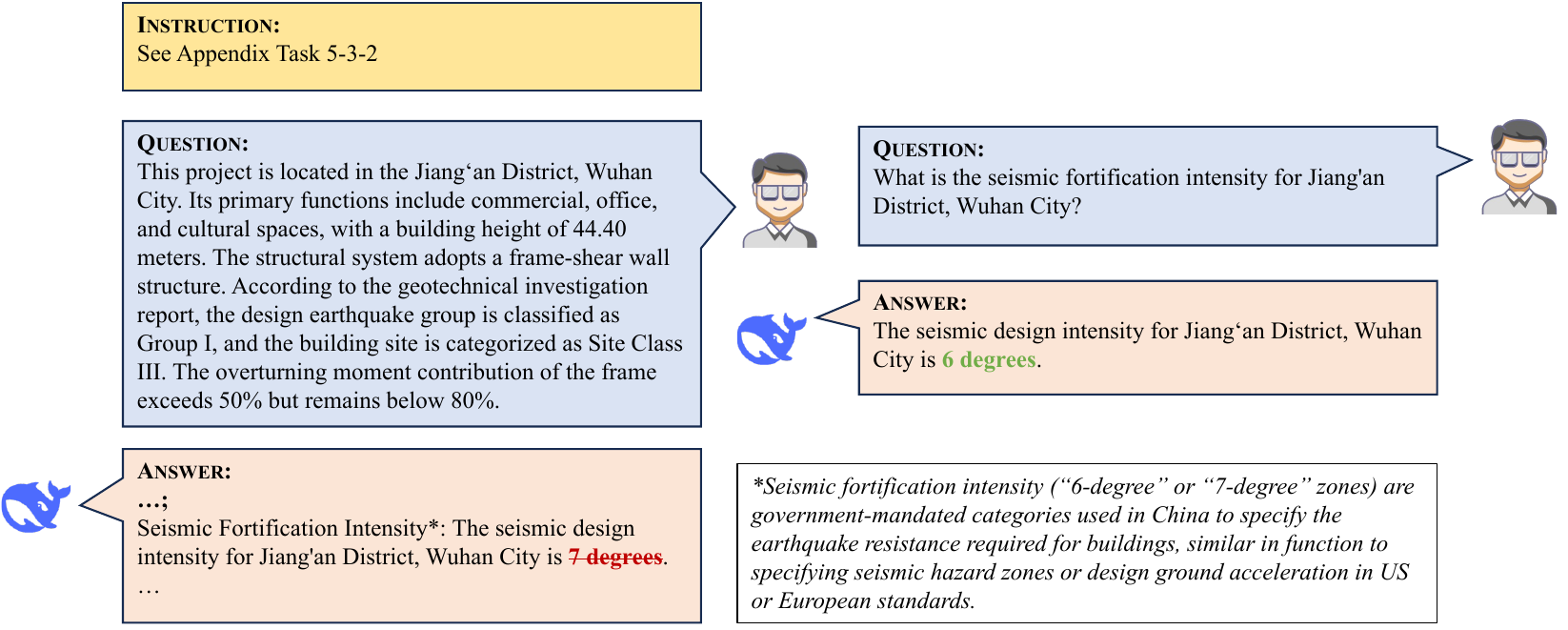}
	\caption{Failure case analysis: the gap between Knowledge Memorization and Application}
	\label{fig:und2app}
\end{figure}

Figure~\ref{fig:und2app} highlights a critical disparity between the model's knowledge memorization and its practical application. The model demonstrated this gap in complex creative tasks; for instance, it consistently failed to determine the correct seismic grade when generating a full design proposal. However, the model could accurately provide the very same seismic grade when asked for it in a direct, isolated query. This proves the model's failure is not a knowledge deficit but an application deficit. While the model may possess the necessary factual knowledge, it cannot reliably access and integrate it into a multi-step reasoning and long-form generation process.

A direct comparison between the DeepSeek-R1 and DeepSeek-V3 is justified by their virtually identical model architectures and parameter scales. The DeepSeek-R1 demonstrates a significant performance advantage over DeepSeek-V3, excelling in 19 of 23 designed tasks. This superiority is most pronounced in tasks at the Knowledge Reasoning and Knowledge Calculation level and the Evaluation and Creation sub-level. This finding suggests that, for models of equivalent scale and architecture, reinforced training that specifically targets reasoning capabilities yields substantial enhancements in handling complex tasks within the AEC sector. For context, other control pairs from the same organizations, such as Qwen-32B vs. Qwen-Turbo and GPT-4o vs. GPT o3-mini, were considered. However, a lack of publicly available data on their internal architectures and parameters precludes a meaningful comparison, and their performance benchmarks do not present a discernible trend.

Among all the models, DeepSeek-R1 exhibits superior performance, standing out as the best-performing model in 14 out of 23 tasks and is validated as the automated judge. The top performance on the remaining nine tasks is dispersed among the other models, and DeepSeek-V3's average performance is second only to DeepSeek-R1. Hunyuan-Turbo-S, the most recently released closed-source model, follows DeepSeek-V3 and outperforms DeepSeek-R1 on four specific tasks. Surprisingly, QwQ-32B, an open-source model with the smallest known parameter size among those tested, performs above average. Its performance is comparable to that of closed-source models such as Qwen-Turbo, GPT-4o, and GPT o3-mini. However, despite DeepSeek-R1's strong and broad capabilities, its performance declines significantly, particularly on the tasks at the Knowledge Application level, a trend observed across all models. The result underscores a broader finding: even the most advanced LLMs still have substantial room for improvement in handling complex AEC-related tasks.

\subsubsection{Specific observations}
At the Knowledge Memorization level, all models performed exceptionally well, with scores universally exceeding 95\% on terminology memorization (Task 1-2). Across these three tasks, model performance exhibited a high mean and low standard deviation. This strongly suggests that a considerable amount of foundational AEC knowledge is already embedded within the parameters of current LLMs. 

At the Knowledge Understanding and Reasoning levels, a notable weakness was observed in tasks requiring the interpretation of tabular data (Tasks 2-3 and 3-3). The performance of the LLMs was notably deficient, even falling below that of the supposedly more advanced application tasks (for example, information extraction tasks). The performance decline is discussed in detail in Section\ref{dis}. Conversely, on the building type inference task (Task 3-2), performance unexpectedly approached the performance of some Knowledge Memorization tasks (for example, code memorization \& retrieval task). An analysis of Task 3-3 indicates that its correct option often exhibits high semantic similarity with the question stem, primarily because of shared keywords. Consequently, successfully completing this task may not strictly require inductive reasoning.

At the Knowledge Calculation level, model performance across the four tasks exhibited a low mean and a high standard deviation. Particularly, the lower scores in architectural design calculation (Task 4-1) and engineering economics calculation (Task 4-4) are attributable to the multi-step analysis and complex computations these problems require, posing a greater challenge to the models. The high standard deviation precisely indicates that this capability is a key differentiator between high-performing models and general models.
At the Knowledge Application level, performance on the sub-levels was highly variable. In the Analysis tasks, models worked well on document classification and proofreading tasks (Tasks 5-1-1 \& 5-1-2) but struggled notably with compliance checking (Task 5-1-3) and brand compliance verification (Task 5-1-4). Lower scores in both Task 5-1-3 and Task 5-1-4 highlight a broader weakness in the ability to accurately compare key information, a core requirement for both tasks. This difficulty is then compounded by the unique demands of each: the former requires complex reasoning and calculation, while the latter necessitates the accurate parsing of tables in Markdown format. In contrast, performance on the information extraction task (Task 5-1-5) showcases the LLMs' proficiency in parsing and extracting information from long-context domain documents, with the Moonshot-v1-128k model achieving the best performance on this task.

In the Evaluation tasks, the modest Kendall's tau coefficients (failing to surpass the 0.7 threshold for strong correlation) are attributable to the profound difficulty of shifting from simple pattern matching to the kind of high-fidelity, evidence-based reasoning process. This process requires LLMs to interpret key evidence in criteria that may be paraphrased, expressed implicitly, or scattered across a long context, posing a significant challenge to the goal of substituting human experts in rubric-based evaluations. The human inter-annotator agreement (IAA) baseline and the subsequent LLM-human consensus correlation are presented in Section~\ref{dis}. In the Creation tasks, all models failed to surpass the 60-point threshold. The uniformly limited performance is attributable to the complex demands of professional document generation. This process requires a multi-faceted synthesis of factual accuracy, logical coherence, and structural norms, posing a significant challenge to the models' advanced generative capabilities.

\section{Discussion}
\label{dis}

\subsection{Systematic bias of LLM-as-a-Judge approach} 

\begin{figure}[!tbp]
  \centering
  \includegraphics[width=1.0\textwidth]{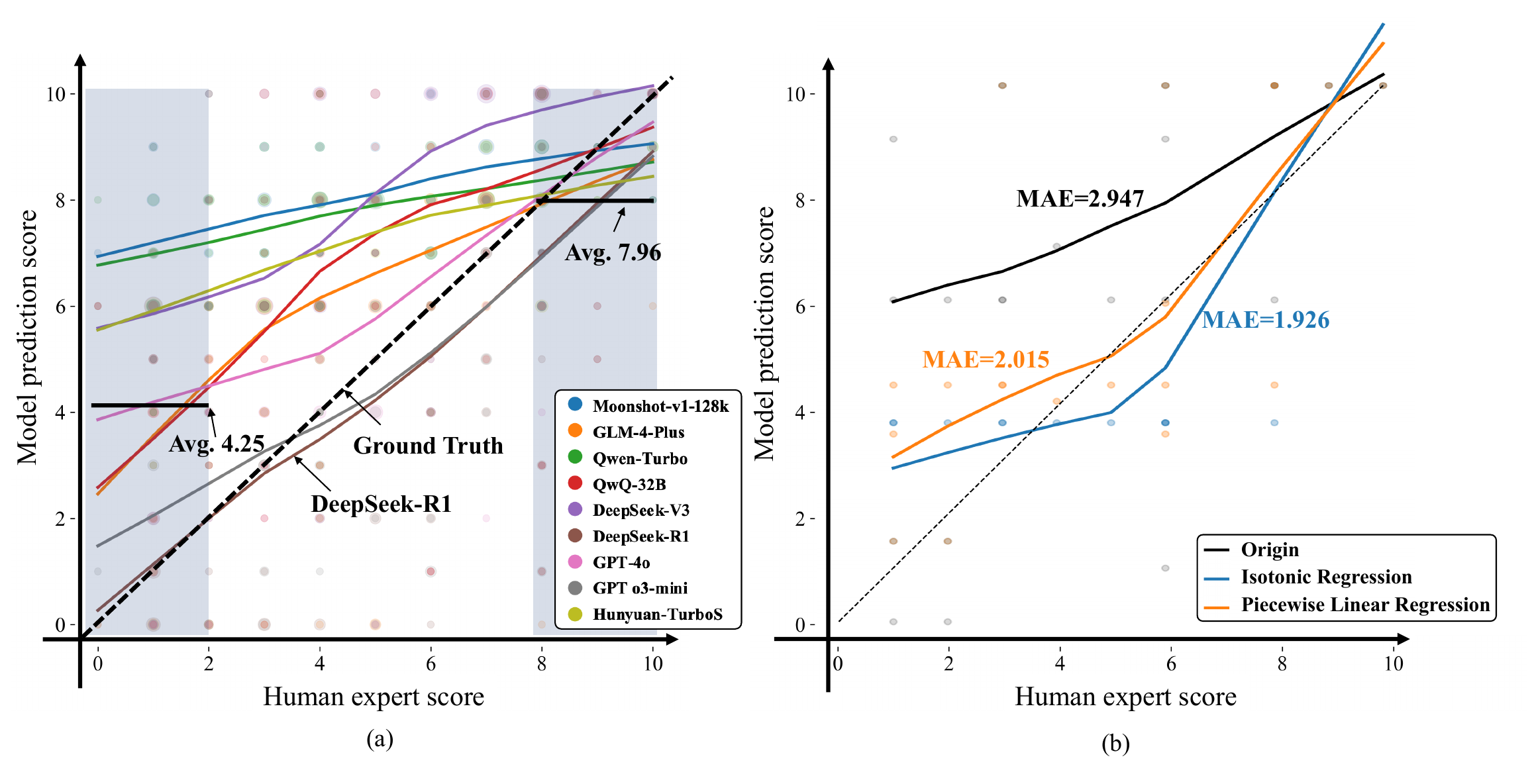}
  \captionsetup{justification=centering}
  \caption{Scatter plots with LOWESS curves of model performance in evaluation tasks\\
           \textit{(a) original performance of LLMs; (b) performance of DeepSeek-V3 with calibration methods}}
  \label{fig:14}
\end{figure}

Figure~\ref{fig:14}(a) presents a scatter plot of LLM prediction scores against human expert scores, where the size of each point is proportional to its frequency of occurrence. A \textit{y=x} line in this figure indicates the ground truth for perfect model-human agreement. For each model, a smoothed curve is fitted to the data points to illustrate its performance trend. While all models demonstrate a generally positive correlation with human assessments, as shown by their monotonically rising curves, DeepSeek-R1’s curve is closest to the \textit{y=x} line, indicating the highest level of agreement. 

However, the plot also reveals a systematic bias: for instance, models tend to assign an average score of 4.25 to documents rated in the 0–2 range by experts, while conversely assigning an average of only 7.96 to those rated between 8 and 10. This tendency to overestimate low scores and underestimate high scores results in a compressed scoring interval. This compression has significant implications for using any model in automated grading, particularly those with less-than-ideal performance. Therefore, to meet the required standards of reliability, it is essential to introduce calibration methods and uncertainty assessments.

Two calibration methods (i.e., isotonic regression and piecewise linear regression) were applied in this paper to mitigate the observed scoring bias~\cite{cali1,cali2}. As illustrated in Figure~\ref{fig:14}(b), both methods successfully shifted the model's performance curve closer to the ideal \textit{y=x} line. This improvement is quantitatively confirmed by a significant reduction in the mean absolute error (MAE) from an initial 2.947 to 1.926 (isotonic regression) and 2.015 (piecewise linear regression). These results demonstrate that both methods effectively reduce the systematic bias of LLM-as-a-Judge, leading to predictions that align more closely with human expert scores.

\subsection{Performance decline in tasks related to code tables}  
\begin{figure}[!tbp]
	\centering
	\includegraphics[width=1.0\textwidth]{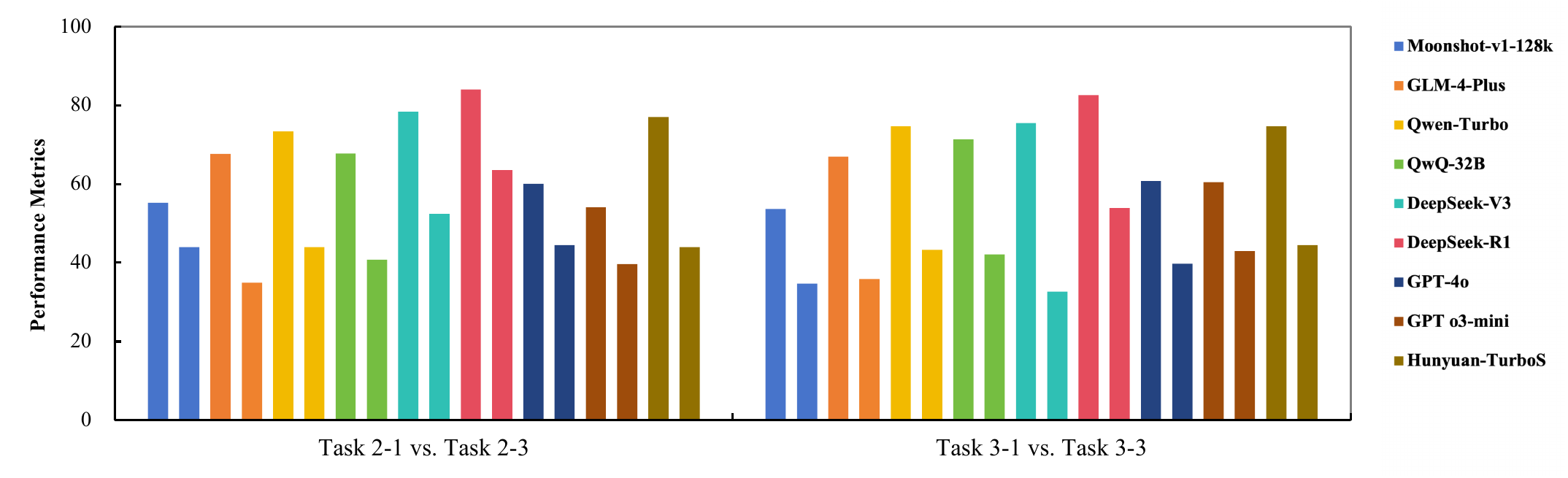}
	\caption{Performance decline in questions related to tables in codes}
	\label{fig:16}
\end{figure}

\begin{figure}[!tbp]
	\centering
	\includegraphics[width=\textwidth]{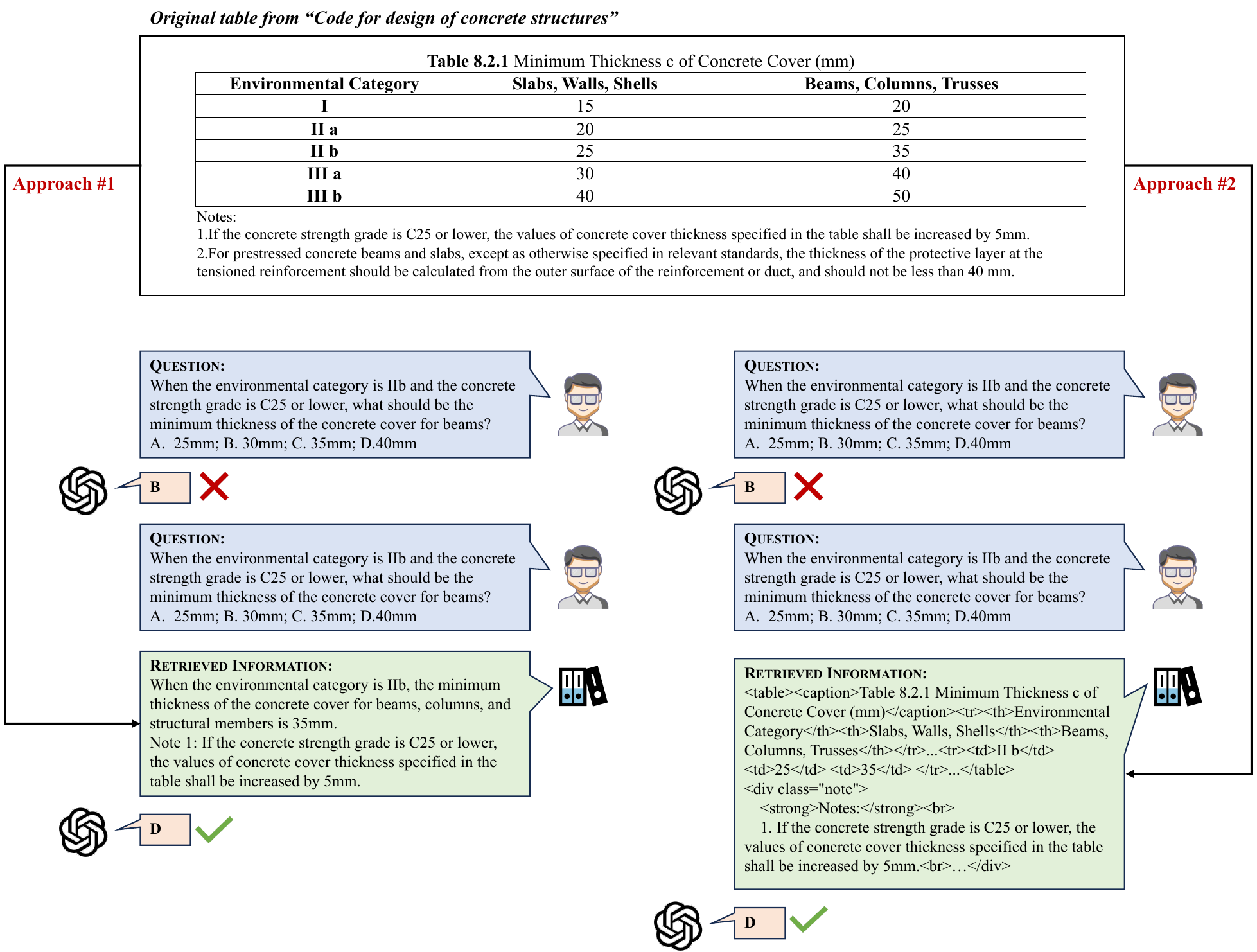}
	\caption{Two context formatting approaches for tables in codes}
	\label{fig:15}
\end{figure}

Figure~\ref{fig:16} presents model performance in two types of comparative tasks designed to specifically assess the model’s ability to understand and reason over knowledge in code tables. A significant and consistent performance degradation on these two tasks could be clearly observed. To investigate the reasons for this decline in consistency, this paper provides the model with the necessary knowledge sources in two distinct approaches. These two approaches enable LLMs to generate responses based on tabular content embedded in codes, as delineated in Figure~\ref{fig:15}. Approach \#1 converts tabular content into natural language context through expert-crafted textual descriptions. Approach \#2 utilizes models with multi-modal capabilities (GPT-4.1 in this paper) to automatically transform image-based tables into HTML formats, which is less intuitive for human interpretation but inherently machine-parsable~\cite{fang2024largelanguagemodelsllmstabular}. Table~\ref{tab:rag_performance} shows the performance of models on these two tasks. Both Approach \#1 and Approach \#2 achieved significant improvements over the baseline results ('Original'). The average accuracies on these two tasks improved from a baseline of 45.27\% and 40.65\% to 98.94\% and 87.27\% with Approach \#1, and to 72.67\% and 68.24\% with Approach \#2, respectively. These significant improvements also suggest that the decline stems from the model's inability to properly encode knowledge from the code tables.

Moreover, this paper compares two effective approaches for transforming knowledge within tables. After being processed by these approaches, the tables can not only serve as context for the model, as shown in Figure~\ref{fig:15}, but also have their knowledge converted into a pre-training dataset for the model. The superior performance of Approach \#1 over Approach \#2 can be attributed to Approach \#2's inability to correctly reproduce the complex layout of tables from building codes, particularly those with spanning cells. Even with advanced multi-modal architectures or dedicated parsing methods, lossless conversion of tabular content remains unattainable~\cite{li2020tablebankbenchmarkdatasettable,chi2019complicatedtablestructurerecognition,paddleocr2020}. This inherent limitation underscores a fundamental trade-off in transforming tabular data for LLMs: while manual, expert-driven methods can achieve high fidelity, they lack scalability; conversely, automated approaches, though rapid and scalable, often struggle to preserve the complete accuracy and integrity of the original content.
 
\begin{table}[!tbp]
\centering
    \footnotesize                 
\setlength\tabcolsep{4pt}   
\renewcommand{\arraystretch}{1.3}
\begin{threeparttable}
\caption{Performance comparison of two context formatting approaches}
\label{tab:rag_performance}
\begin{tabular}{lcccccc}
\toprule
\multirow{2}{*}{Model}& \multicolumn{3}{c}{Task 2-3} & \multicolumn{3}{c}{Task 3-3} \\
\cmidrule(lr){2-4} \cmidrule(lr){5-7}
 & Original & Appr. 1 & Appr. 2 & Original & Appr. 1 & Appr. 2 \\
\midrule
Moonshot-v1-128k & 43.92 & 97.88(+53.96) & 71.96(+28.04) & 34.65 & 80.71(+46.06) & 62.99(+28.34) \\
GLM-4-Plus & 34.92 & 100(+65.08) & 70.9(+35.98) & 35.83 & 87.01(+51.18) & 66.54(+30.71) \\
Qwen-Turbo & 43.92 & 98.41(+54.49) & 68.78(+24.86) & 43.31 & 83.07(+39.76) & 64.57(+21.26) \\
QwQ-32B & 40.74 & 98.41(+57.67) & 73.54(+32.8) & 42.13 & 88.19(+46.06) & 71.65(+29.52) \\
DeepSeek-V3 & 52.38 & 100(+47.62) & 68.78(+16.4) & 32.68 & 84.25(+51.57) & 60.24(+27.56) \\
DeepSeek-R1 & 63.49 & 99.47(+35.98) & 76.72(+13.23) & 53.94 & 90.94(+37) & 74.41(+20.47) \\
GPT-4o & 44.44 & 97.88(+53.44) & 74.6(+30.16) & 39.76 & 91.73(+51.97) & 72.83(+33.07) \\
GPT o3-mini & 39.68 & 100(+60.32) & 75.66(+35.98) & 42.91 & 90.55(+47.64) & 70.87(+27.96) \\
Hunyuan-TurboS & 43.92 & 98.41(+54.49) & 73.02(+29.1) & 44.49 & 88.98(+44.49) & 70.08(+25.59) \\
\bottomrule
\end{tabular}
\end{threeparttable}
\end{table}

\label{sec:5}
\section{Conclusion}
\label{sec:6}
To guide the responsible integration and future development of large language models (LLMs) in augmenting the complex, knowledge-intensive tasks of the Architecture, Engineering, and Construction (AEC) industry, this paper established a foundational evaluation benchmark, AECBench, to assess the strengths and limitations of current LLMs in these contexts. A five-level hierarchical framework (\textit{Knowledge Memorization, Knowledge Understanding, Knowledge Reasoning, Knowledge Calculation}, and \textit{Knowledge Application}) was established to mirror domain-specific cognitive demands. A high-quality dataset of 4,800 questions across 23 tasks was primarily created by engineers and subsequently validated by domain experts through a two-round review process. A scalable, LLM-driven evaluation pipeline guided by expert-defined rubrics was developed to assess the complex and long-form responses to open-ended questions. 

The evaluation of nine LLMs revealed a clear performance trend of LLMs in AEC applications. While the models demonstrate exceptional proficiency in memorizing domain-specific knowledge, their performance consistently degrades as the cognitive complexity of the tasks increases. This trend reveals several fundamental limitations of current LLMs in the AEC domain: (i) a notable deficiency in complex reasoning and calculation; (ii) a modest correlation with human expert judgments in rubric-based evaluations, and (iii) a lack of factual accuracy in professional document generation. DeepSeek-R1 demonstrates strong performance, consistently placing it among the top-performing models in \bench{}.

The models' capabilities as rubric-based judges were thoroughly discussed in this paper. The analysis uncovers a systematic bias, and in response, two effective calibration methods (i.e., isotonic regression and piecewise linear regression) were proposed to enhance their reliability. The study also discussed the models' unexpectedly low performance on tasks involving tabular content from building codes. The low performance reveals another limitation about an insufficient mastery of the knowledge within building code tables. While two viable context formatting approaches (i.e., manual textual description and automated HTML conversion) were established, the challenge of efficiently transforming tabular data into fully utilizable and correct content for LLMs remains a critical area for future work.

Future work would incorporate tasks related to multimodal data, such as architectural drawing recognition, to better reflect the full spectrum of information modalities in the AEC field. It is noted that AECBench will be made open-source shortly afterwards. The community is invited to contribute to the enhancement.

\section*{Acknowledgment}
\label{sec:9}
  This project is supported by Shanghai Qi Zhi Institute (Grant No.SQZ202309) and the Scientific Research Project Fund of Arcplus Group PLC (Grant No.24-1L-0142-Z). We extend our gratitude to every engineer and researcher who contributed to this dataset.

\printbibliography

\clearpage
\appendix
\section{Details of tasks}
\label{app1}

\subsection{Knowledge Memorization}

\begin{center}
\begin{minipage}{\textwidth}
	\centering
	\includegraphics[width=0.95\textwidth]{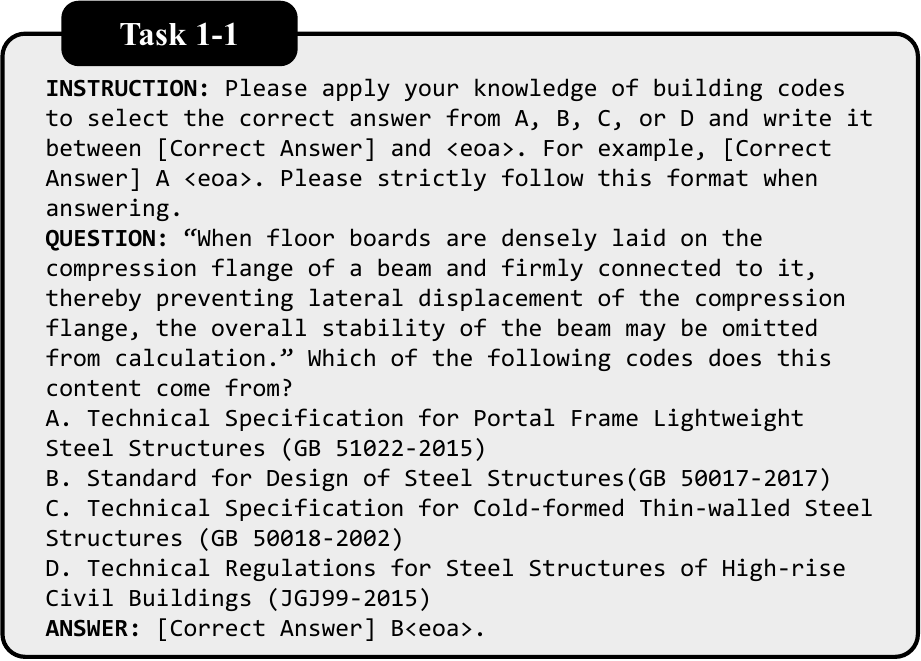}
	\captionof{figure}{The instruction and an example of Task 1-1 Code Memorization \& Retrieval}
	\label{fig:1-1c}

	\vspace{10pt}

	\includegraphics[width=0.95\textwidth]{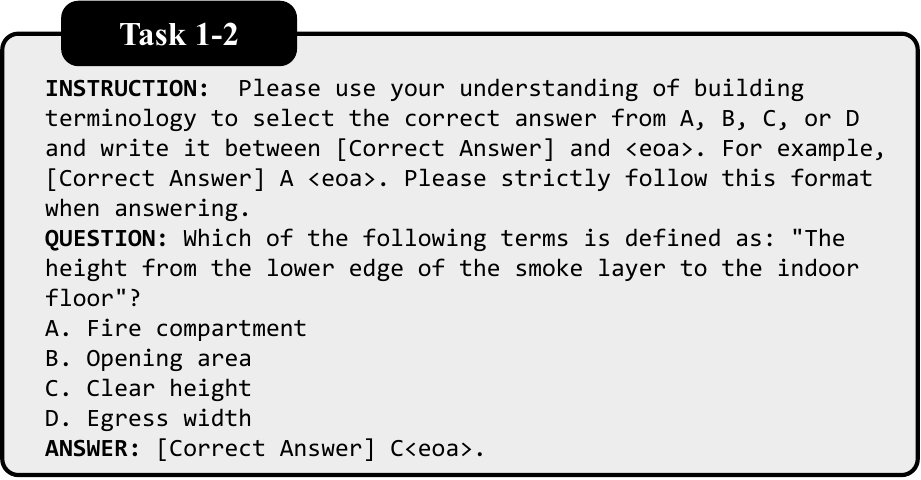}
	\captionof{figure}{The instruction and an example of Task 1-2 Terminology Memorization}
	\label{fig:1-2c}
\end{minipage}
\end{center}

\clearpage
\begin{figure}[htbp]
	\centering
	\includegraphics[width=0.95\textwidth]{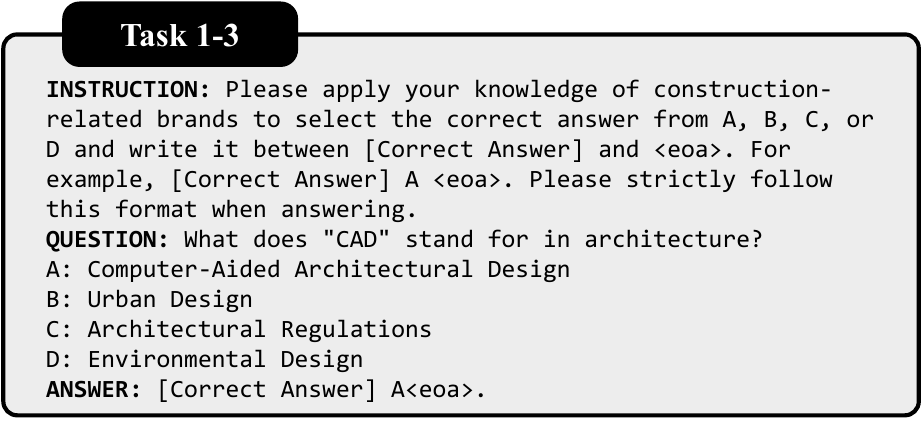}
	\caption{The instruction and an example of Task 1-3 Abbreviation Memorization}
	\label{fig:1-3c}
\end{figure}

\clearpage
\subsection{Knowledge Understanding}

\begin{figure}[htbp]
	\centering
	\includegraphics[width=0.95\textwidth]{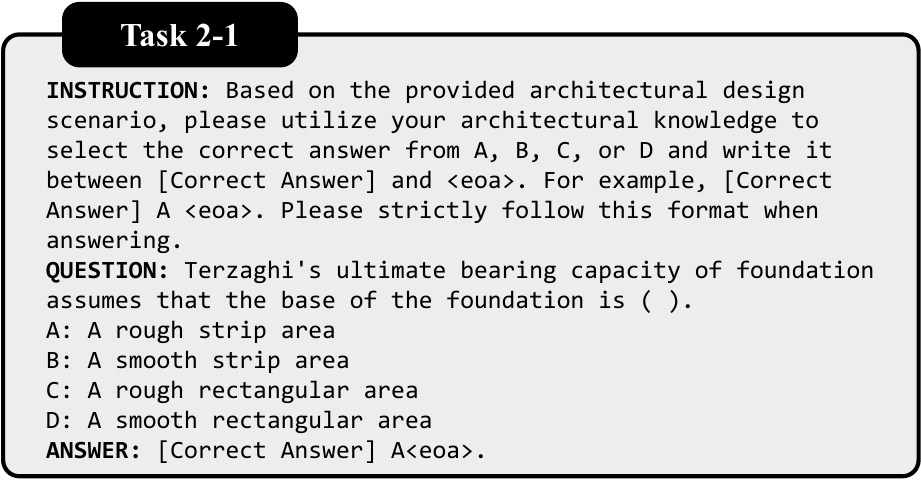}
	\caption{The instruction and an example of Task 2-1 Code Provision Interpretation}
	\label{fig:2-1c}
\end{figure}

\vspace{10pt} 

\begin{figure}[htbp]
	\centering
	\includegraphics[width=0.95\textwidth]{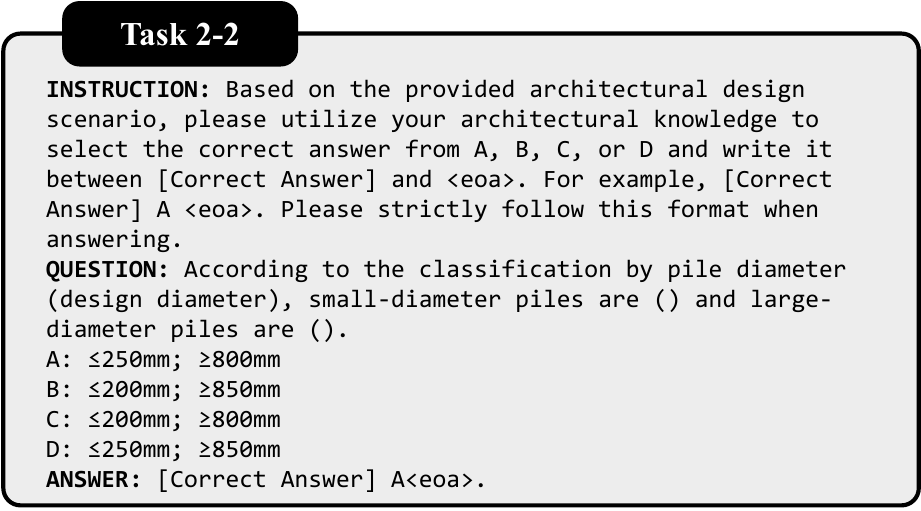}
	\caption{The instruction and an example of Task 2-2 Design Literacy Q \& A}
	\label{fig:2-2c}
\end{figure}

\clearpage

\begin{figure}[htbp]
	\centering
	\includegraphics[width=0.95\textwidth]{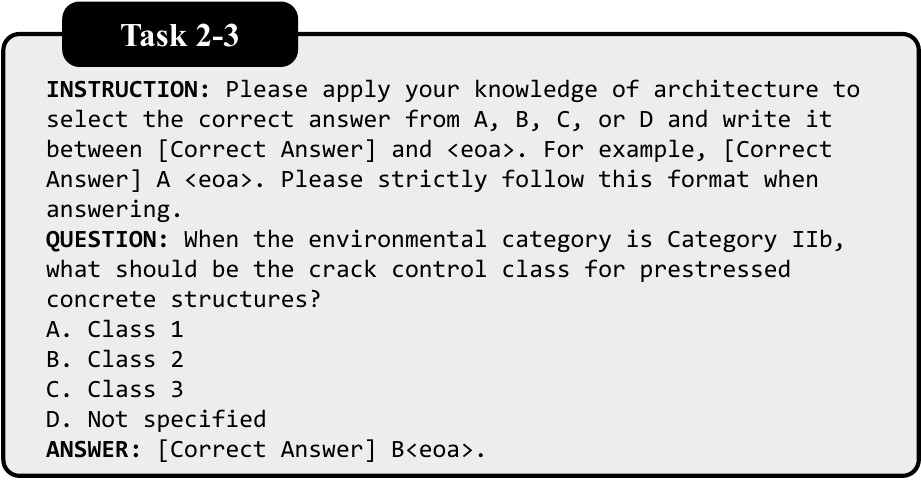}
	\caption{The instruction and an example of Task 2-3 Code Provision Interpretation(Tabular Data)}
	\label{fig:2-4c}
\end{figure}
\clearpage
\subsection{Knowledge Reasoning}
\vspace{-5pt}  
\begin{figure}[htbp]
	\centering
	\includegraphics[width=0.95\textwidth]{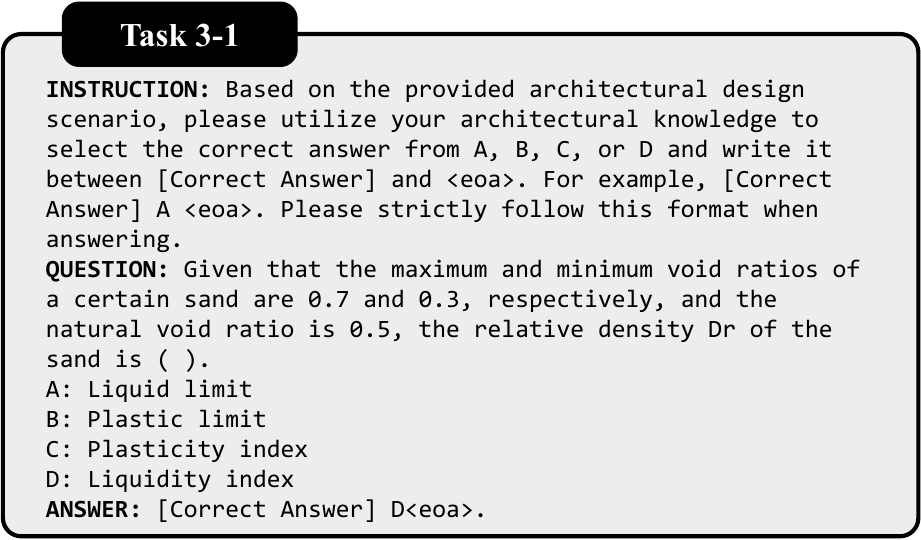}
	\caption{The instruction and an example of Task 3-1 Design Decision Formulation}
	\label{fig:3-1c}
\end{figure}
\vspace{10pt} 
\begin{figure}[htbp]
	\centering
	\includegraphics[width=0.95\textwidth]{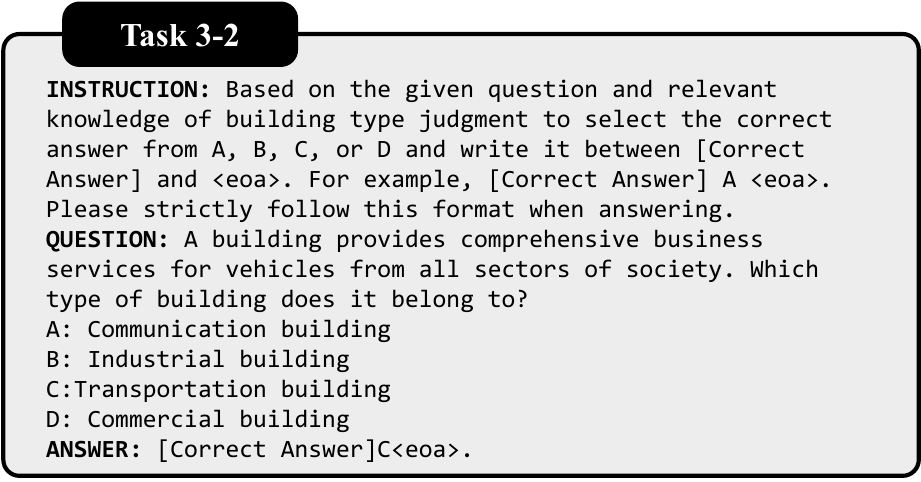}
	\caption{The instruction and an example of Task 3-2 Building Type Prediction}
	\label{fig:3-2c}
\end{figure}

\clearpage
\begin{figure}[htbp]
	\centering
	\includegraphics[width=0.95\textwidth]{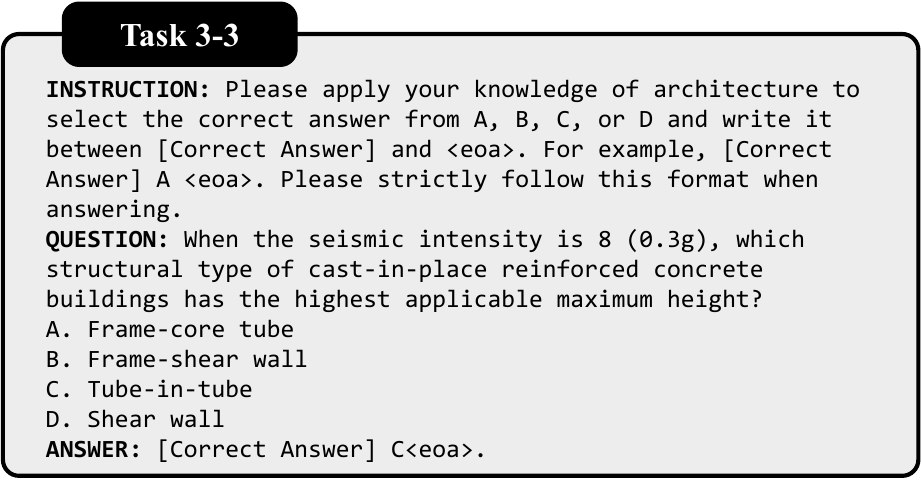}
	\caption{The instruction and an example of Task 3-3 Design Decision Formulation(Tabular Data)}
	\label{fig:3-3c}
\end{figure}
\clearpage
\subsection{Knowledge Calculation}
\begin{figure}[htbp]
	\centering
	\includegraphics[width=0.95\textwidth]{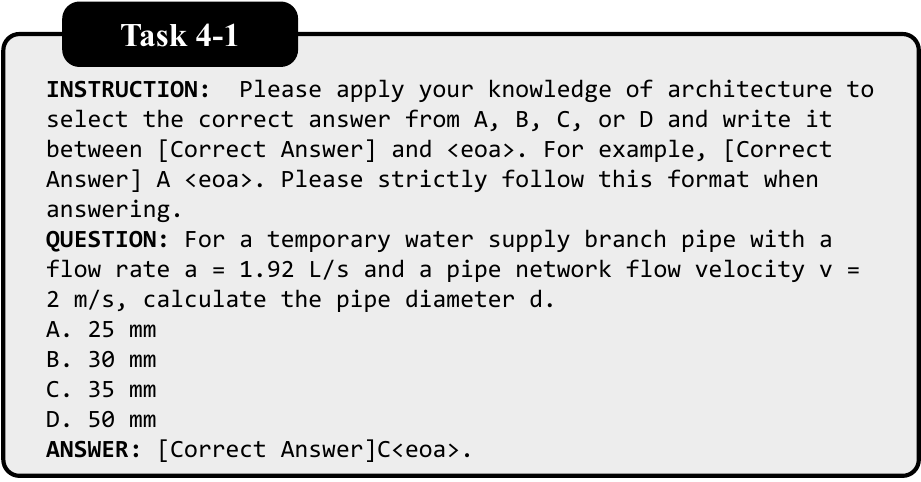}
	\caption{The instruction and an example of Task 4-1 Calculation in Architectural Design}
	\label{fig:4-1c}
\end{figure}

\begin{figure}[htbp]
	\centering
	\includegraphics[width=0.95\textwidth]{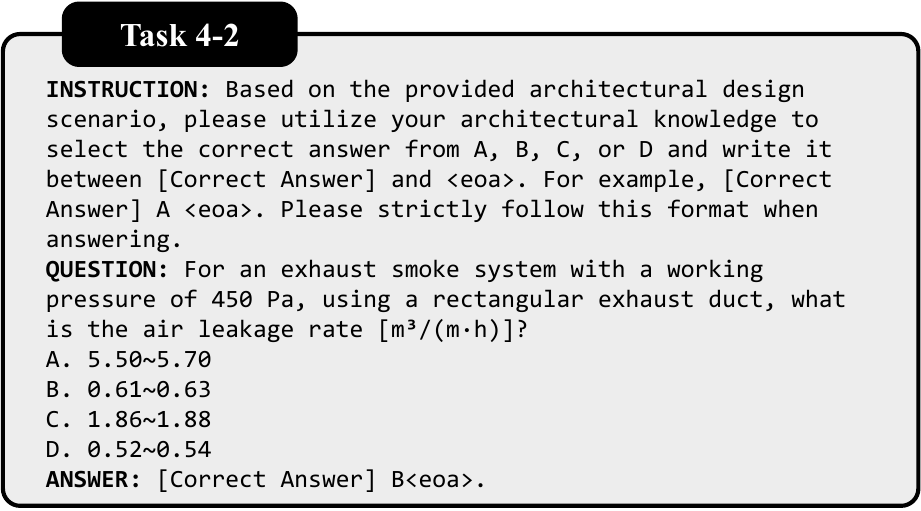}
	\caption{The instruction and an example of Task 4-2 Calculation in MEP Design}
	\label{fig:4-2c}
\end{figure}

\begin{figure}[htbp]
	\centering
	\includegraphics[width=0.95\textwidth]{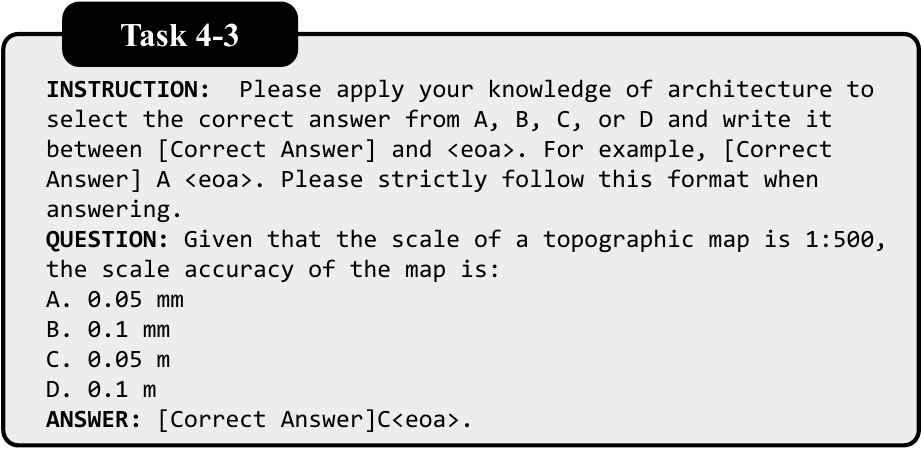}
	\caption{The instruction and an example of Task 4-3 Calculation in Construction}
	\label{fig:4-3c}
\end{figure}

\begin{figure}[!tbp]
	\centering
	\includegraphics[width=0.95\textwidth]{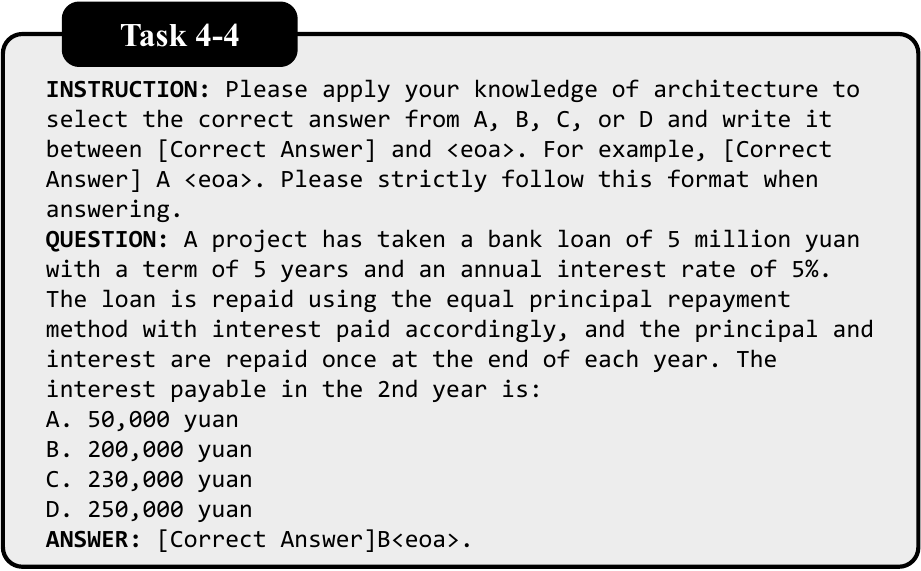}
	\caption{The instruction and an example of Task 4-4 Calculation in Engineering Economics}
	\label{fig:4-4c}
\end{figure}
\clearpage
\subsection{Knowledge Application}
\vspace{0pt}

\begin{center}
\begin{minipage}{\textwidth}
	\centering
	\includegraphics[width=0.95\textwidth]{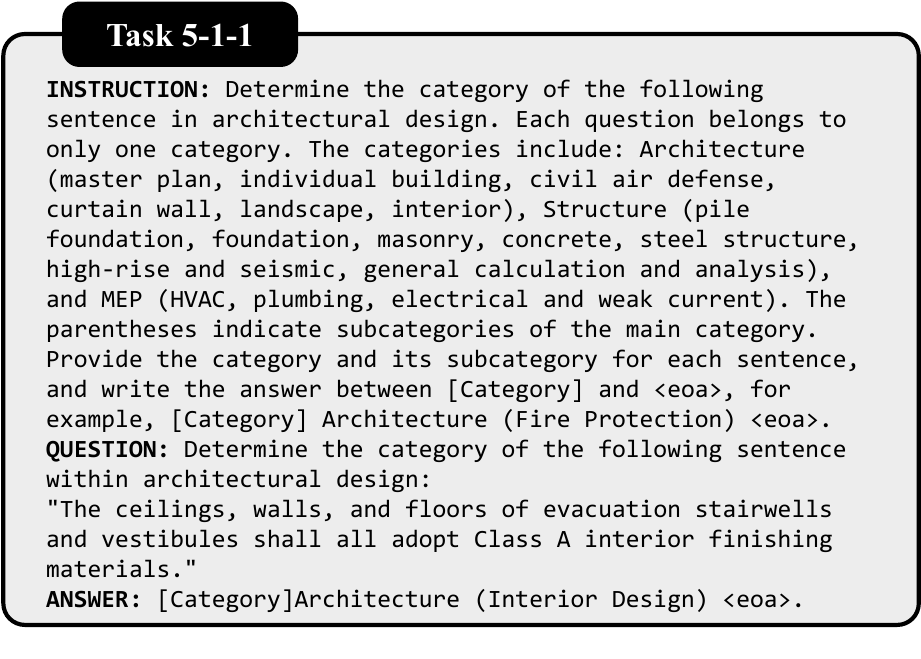}
	\captionof{figure}{The instruction and an example of Task 5-1-1 Document Classification}
	\label{fig:5-1-1c}

	\vspace{10pt}

	\includegraphics[width=0.95\textwidth]{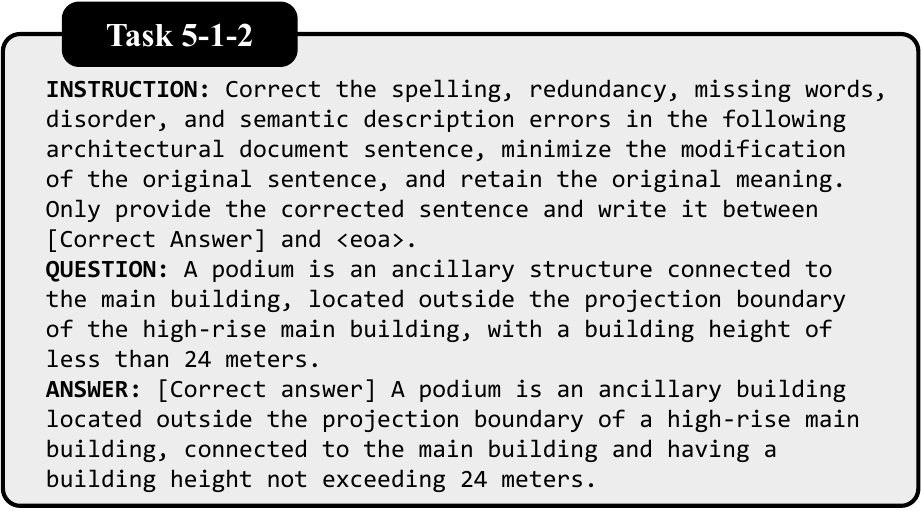}
	\captionof{figure}{The instruction and an example of Task 5-1-2 Document Proofreading}
	\label{fig:5-1-2c}
\end{minipage}
\end{center}

\clearpage
\begin{figure}[htbp]
	\centering
	\includegraphics[width=0.95\textwidth]{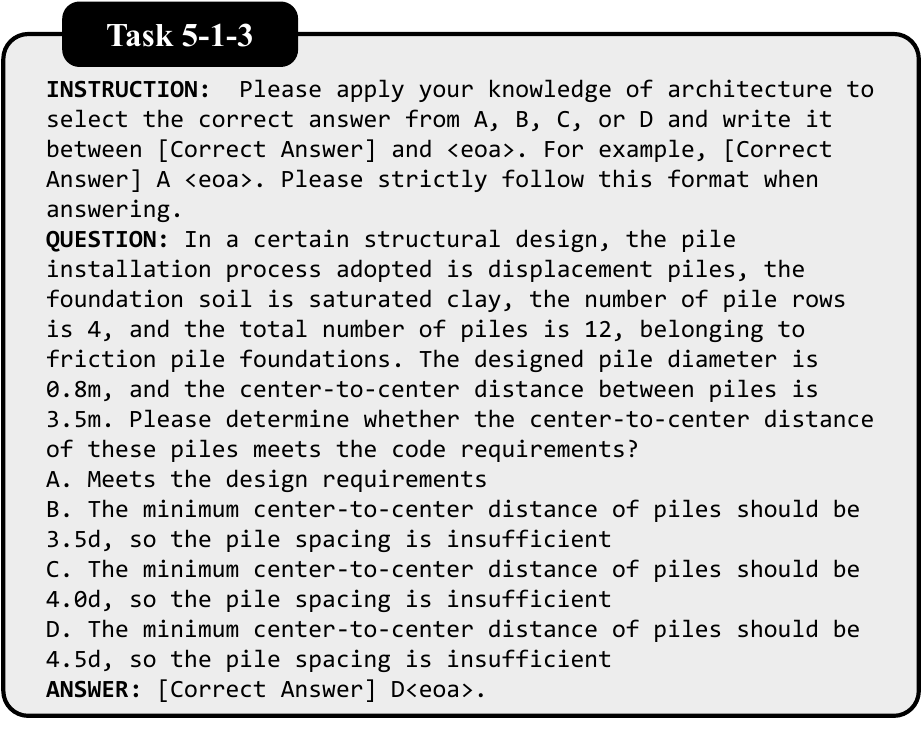}
	\vspace{-10pt}
	\caption{The instruction and an example of Task 5-1-3 Compliance Checking}
	\label{fig:5-1-3c}
\end{figure}
\clearpage
\begin{figure}[htbp]
	\centering
	\includegraphics[width=0.95\textwidth]{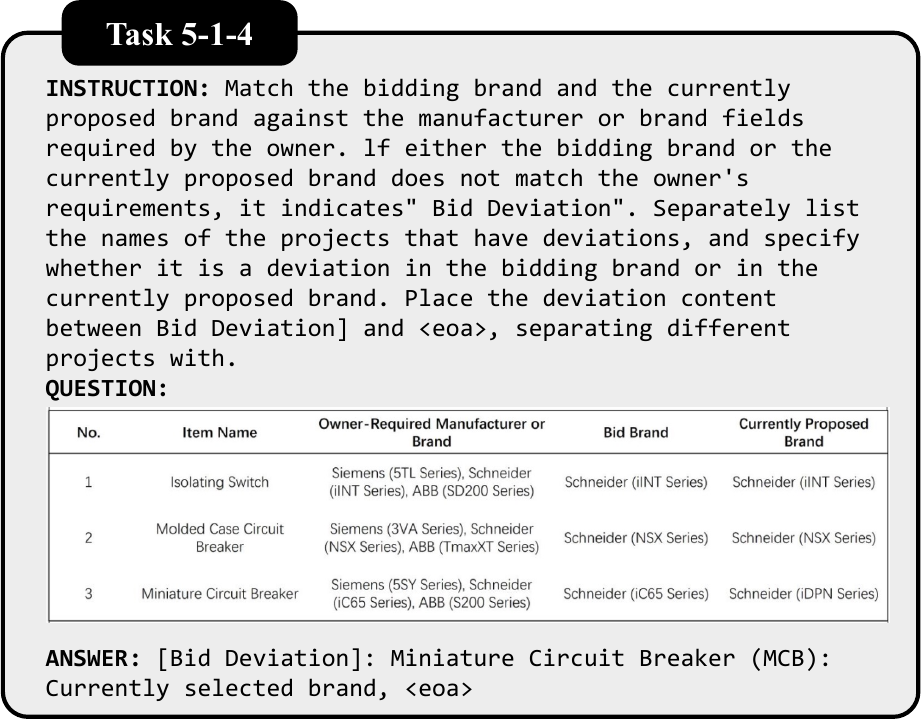}
	\caption{The instruction and an example of Task 5-1-4 Brand Compliance Verification}
	\label{fig:5-1-4c1}
\end{figure}

\clearpage
\begin{figure}[p]   
	\vspace{-20pt}
	\centering
	\includegraphics[width=0.95\textwidth]{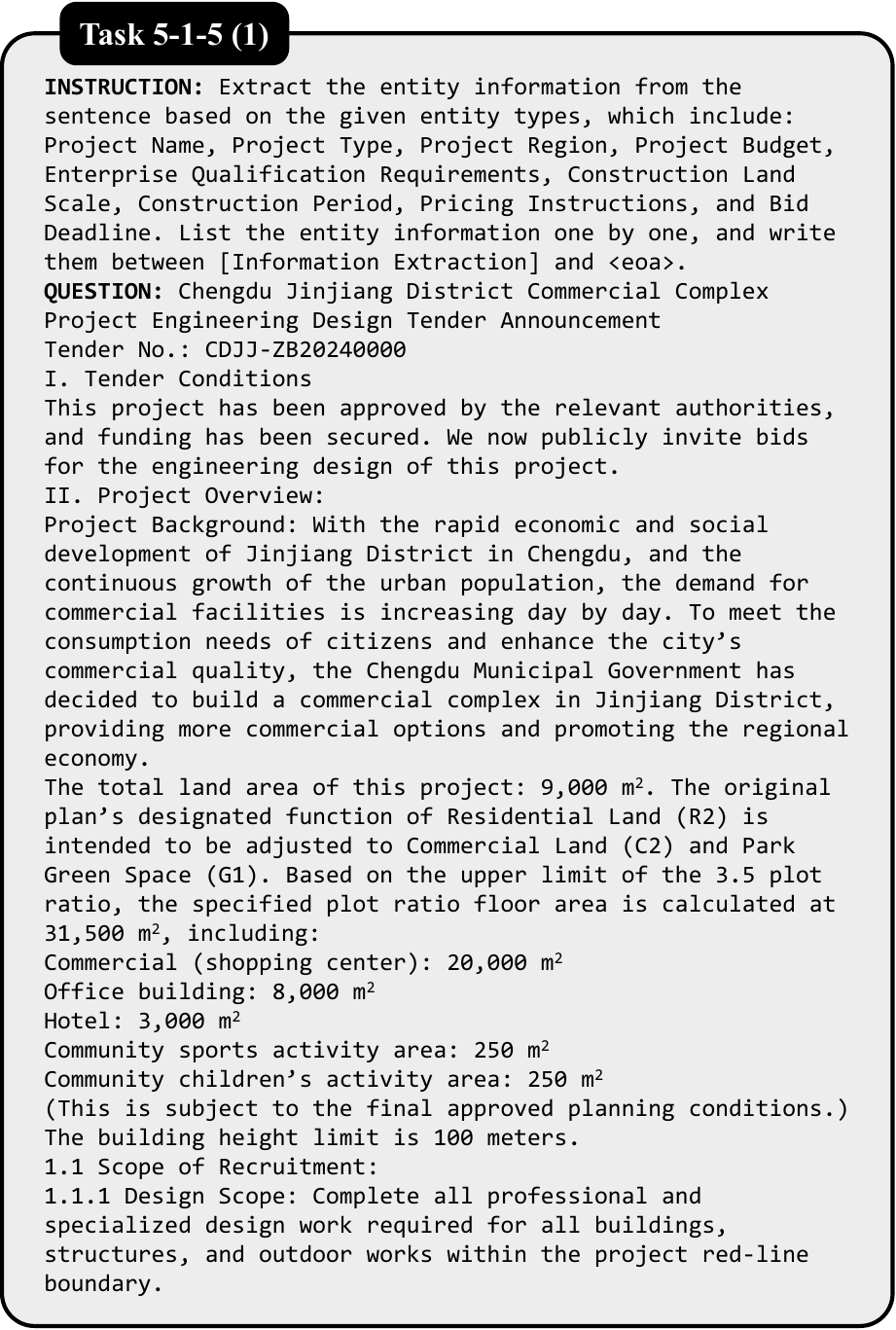}
	\caption{The instruction and an example of Task 5-1-5 Information Extraction \\(continued on thenext page)}
	\label{fig:5-1-5c2}
\end{figure}

\clearpage
\begin{figure}[p]
	\ContinuedFloat
	\vspace{-20pt}
	\centering
	\includegraphics[width=0.95\textwidth]{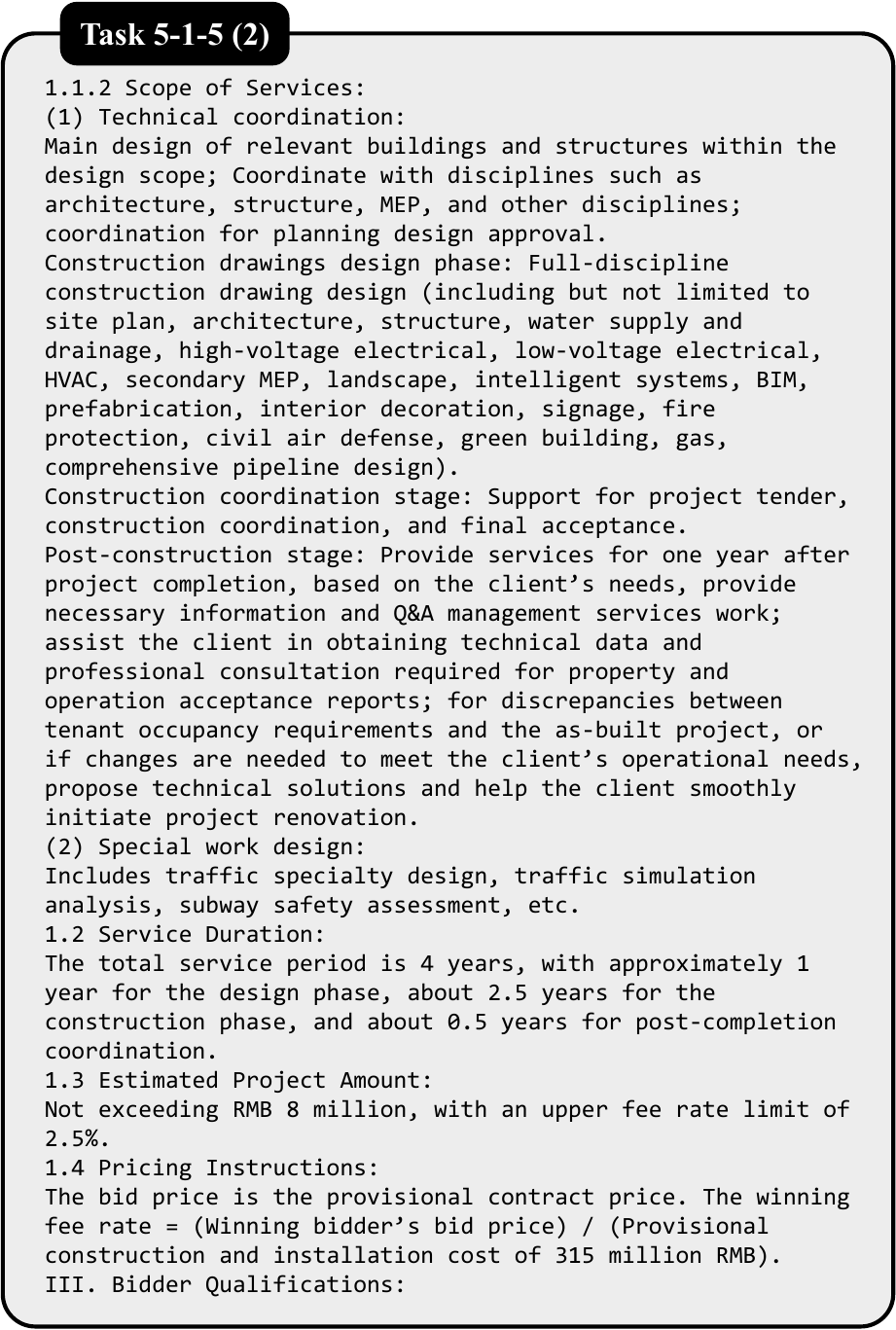}
	\caption*{Figure \thefigure: The instruction and an example of Task 5-1-5 Information Extraction (continued)}
	\label{fig:5-1-5c3}
\end{figure}

\clearpage
\begin{figure}[p]
	\ContinuedFloat
	\vspace{-20pt}
	\centering
	\includegraphics[width=0.95\textwidth]{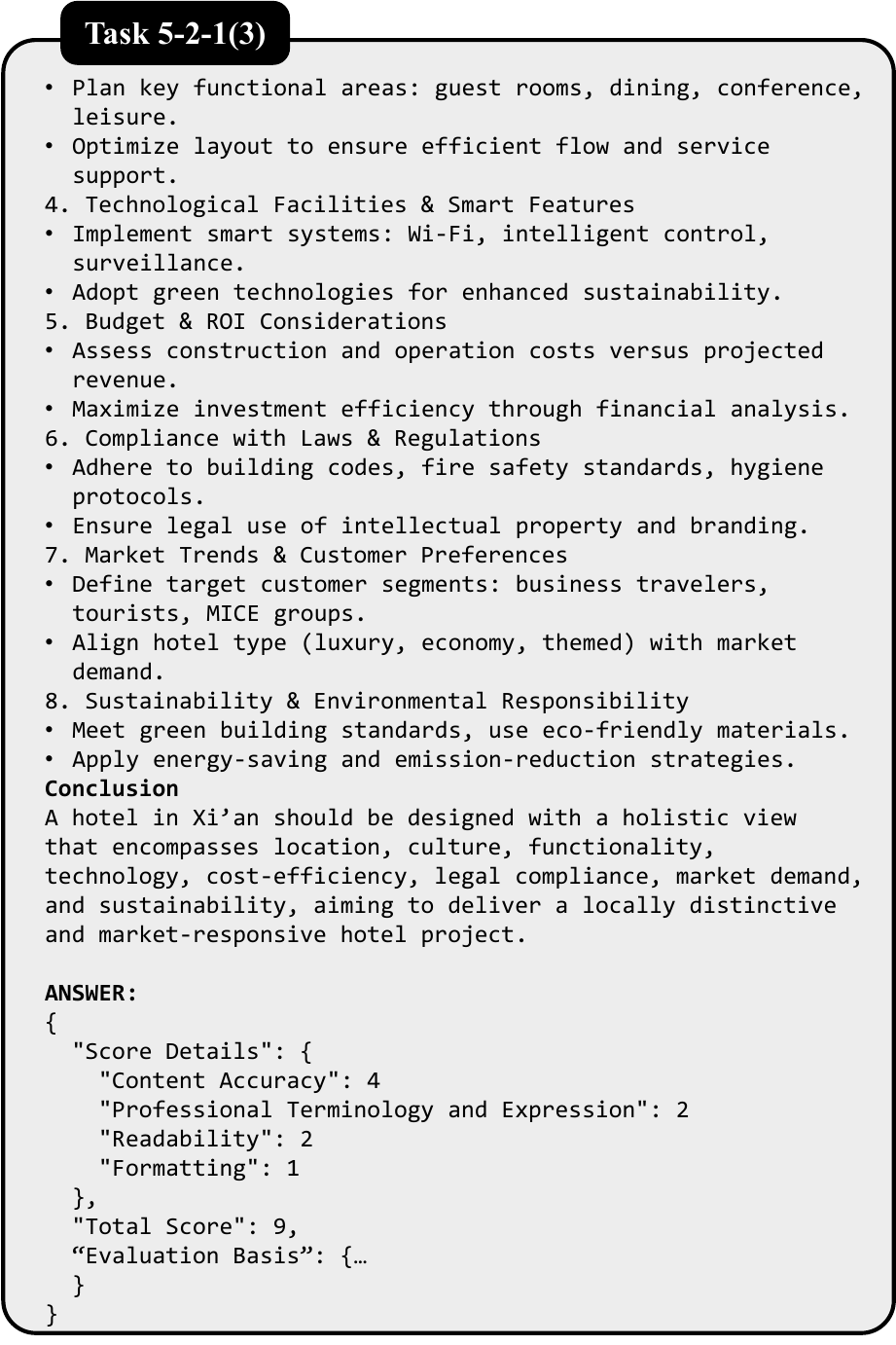}    \caption*{Figure \thefigure: The instruction and an example of Task 5-1-5 Information Extraction (continued)}
	\label{fig:5-1-5c4}
\end{figure}

\clearpage
\begin{figure}[htbp]
	\centering
	\includegraphics[width=0.95\textwidth]{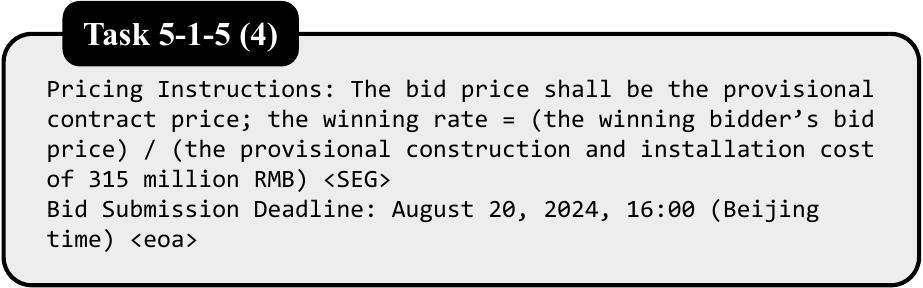}    \caption*{Figure \thefigure: The instruction and an example of Task 5-1-5 Information Extraction (continued)}
	\label{fig:5-1-5c5}
\end{figure}

\clearpage
\begin{figure}[p]   
	\vspace{-20pt}
	\centering
	\includegraphics[width=0.95\textwidth]{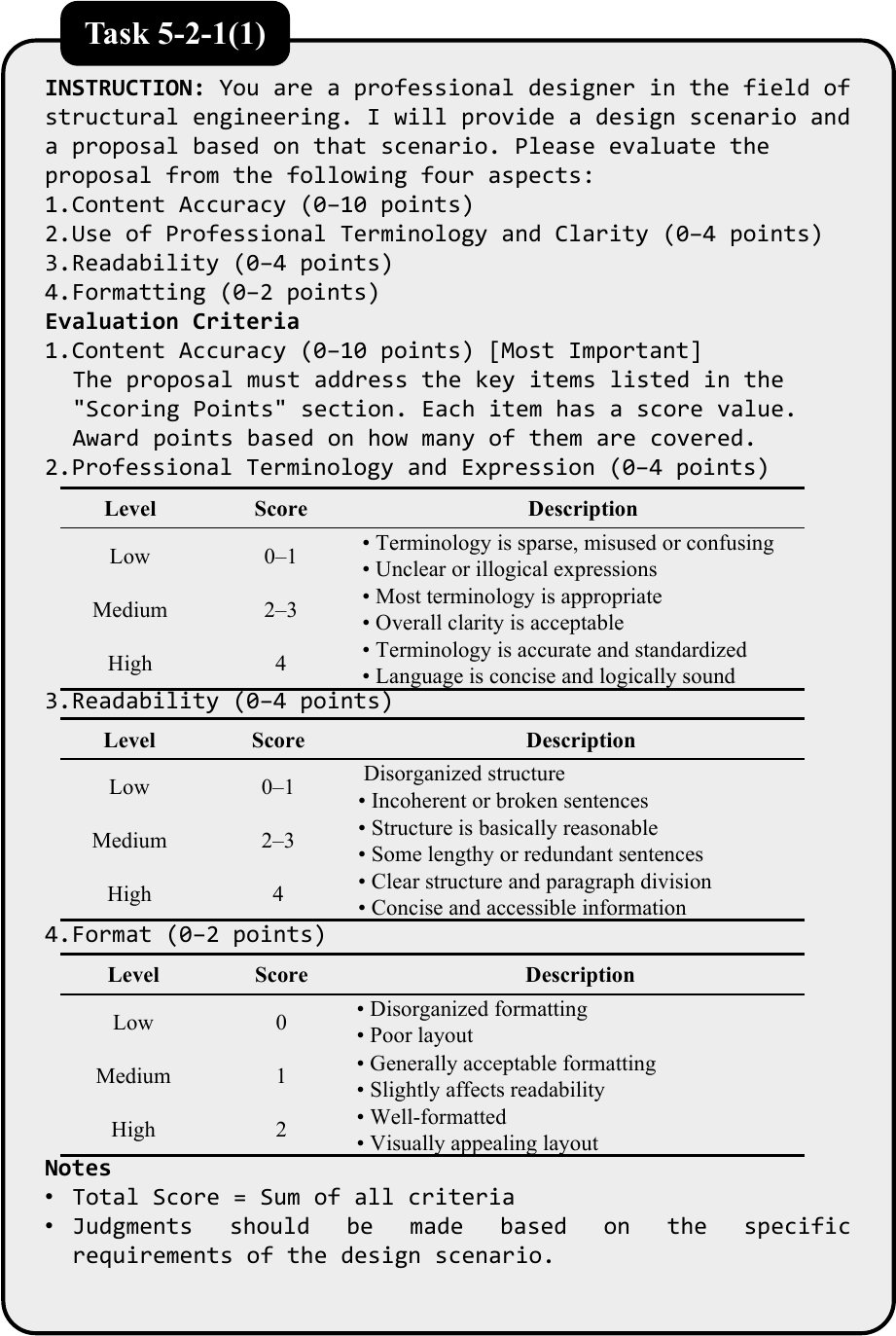}
	\caption{The instruction and an example of Task 5-2-1 Evaluation of Documents in Architectural Design\\(continued on the next page)}
	\label{fig:5-2-1c1}
\end{figure}

\clearpage
\begin{figure}[p]
	\ContinuedFloat
	\vspace{-20pt}
	\centering
	\includegraphics[width=0.95\textwidth]{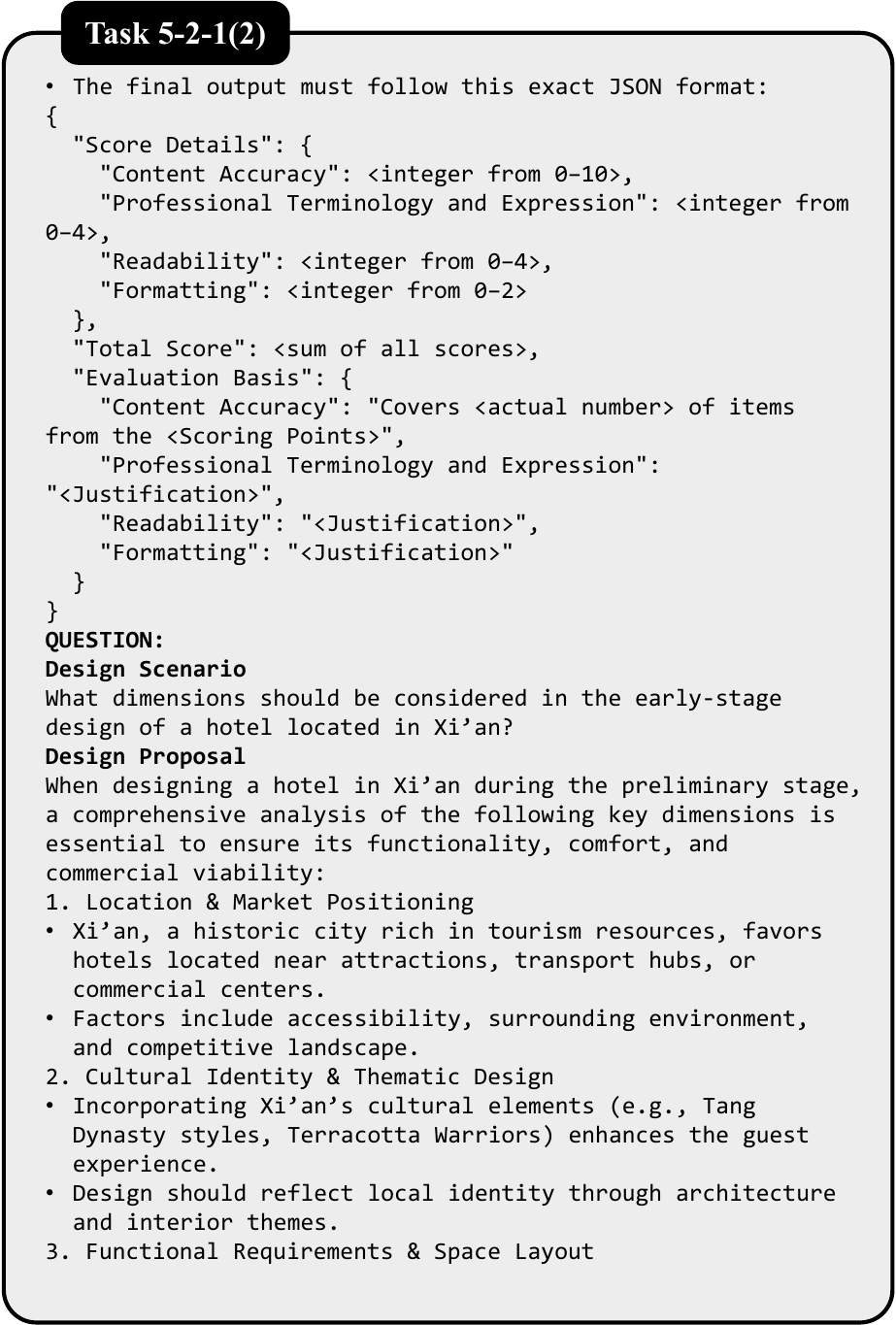}
	\caption*{Figure \thefigure: The instruction and an example of Task 5-2-1 Evaluation of Documents in Architectural Design(continued)}
	\label{fig:5-2-1c2}
\end{figure}

\clearpage
\begin{figure}[p]
	\ContinuedFloat
	\vspace{-20pt}
	\centering
	\includegraphics[width=0.95\textwidth]{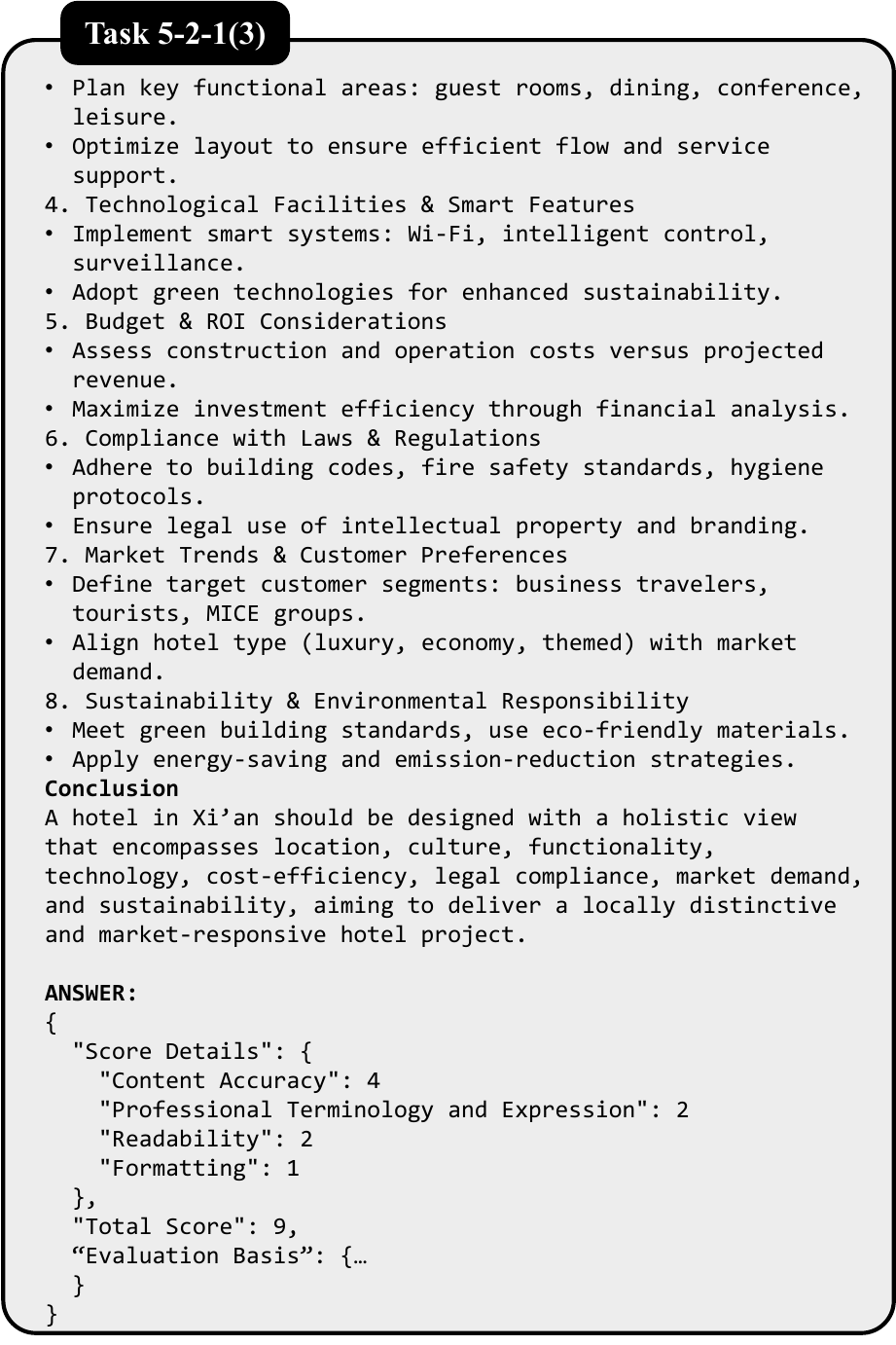}    \caption*{Figure \thefigure: The instruction and an example of Task 5-2-1 Evaluation of Documents in Architectural Design(continued)}
	\label{fig:5-2-1c3}
\end{figure}

\clearpage
\begin{figure}[p]   
	\vspace{-20pt}
	\centering
	\includegraphics[width=0.95\textwidth]{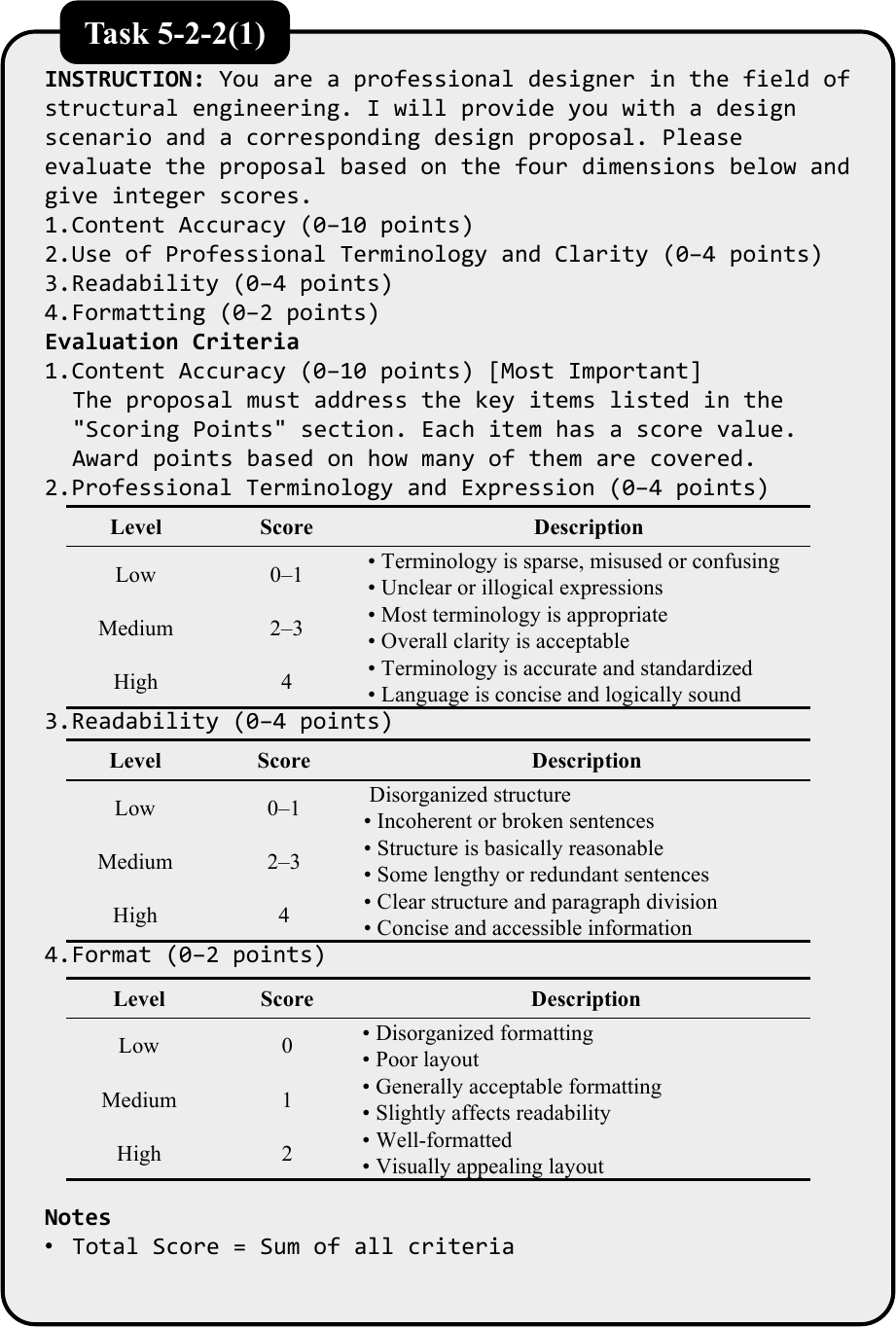}
	\caption{The instruction and an example of Task 5-2-2 Evaluation of Documents in Structural Design\\(continued on the next page)}
	\label{fig:5-2-2c1}
\end{figure}

\clearpage
\begin{figure}[p]
	\ContinuedFloat
	\vspace{-20pt}
	\centering
	\includegraphics[width=0.95\textwidth]{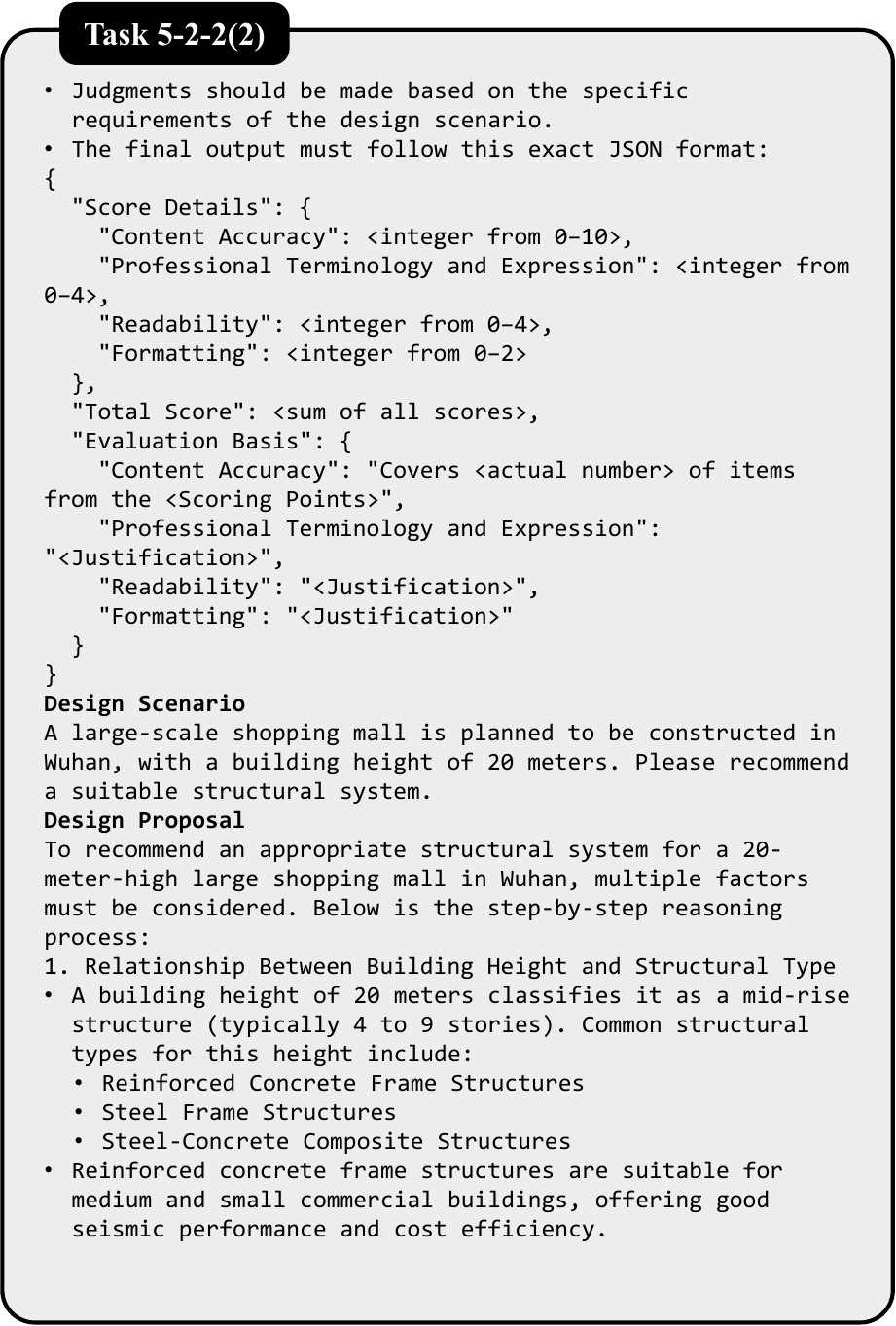}
	\caption*{Figure \thefigure: The instruction and an example of Task 5-2-2 Evaluation of Documents in Structural Design(continued)}
	\label{fig:5-2-2c2}
\end{figure}

\clearpage
\begin{figure}[p]
	\ContinuedFloat
	\vspace{-20pt}
	\centering
	\includegraphics[width=0.95\textwidth]{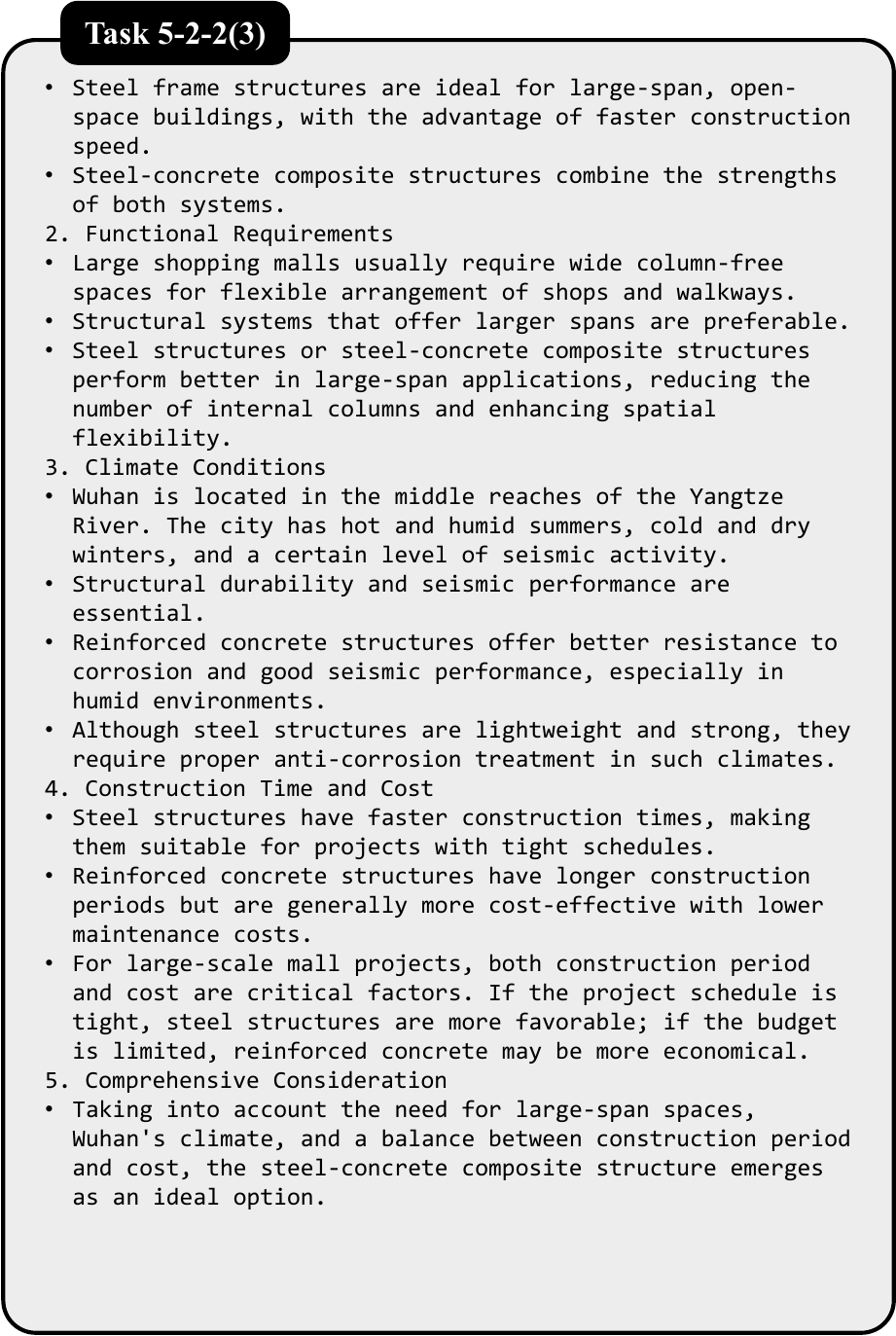}    \caption*{Figure \thefigure: The instruction and an example of Task 5-2-2 Evaluation of Documents in Structural Design(continued)}
	\label{fig:5-2-2c3}
\end{figure}

\clearpage
\begin{figure}[p]
	\ContinuedFloat
	\vspace{-20pt}
	\centering
	\includegraphics[width=0.95\textwidth]{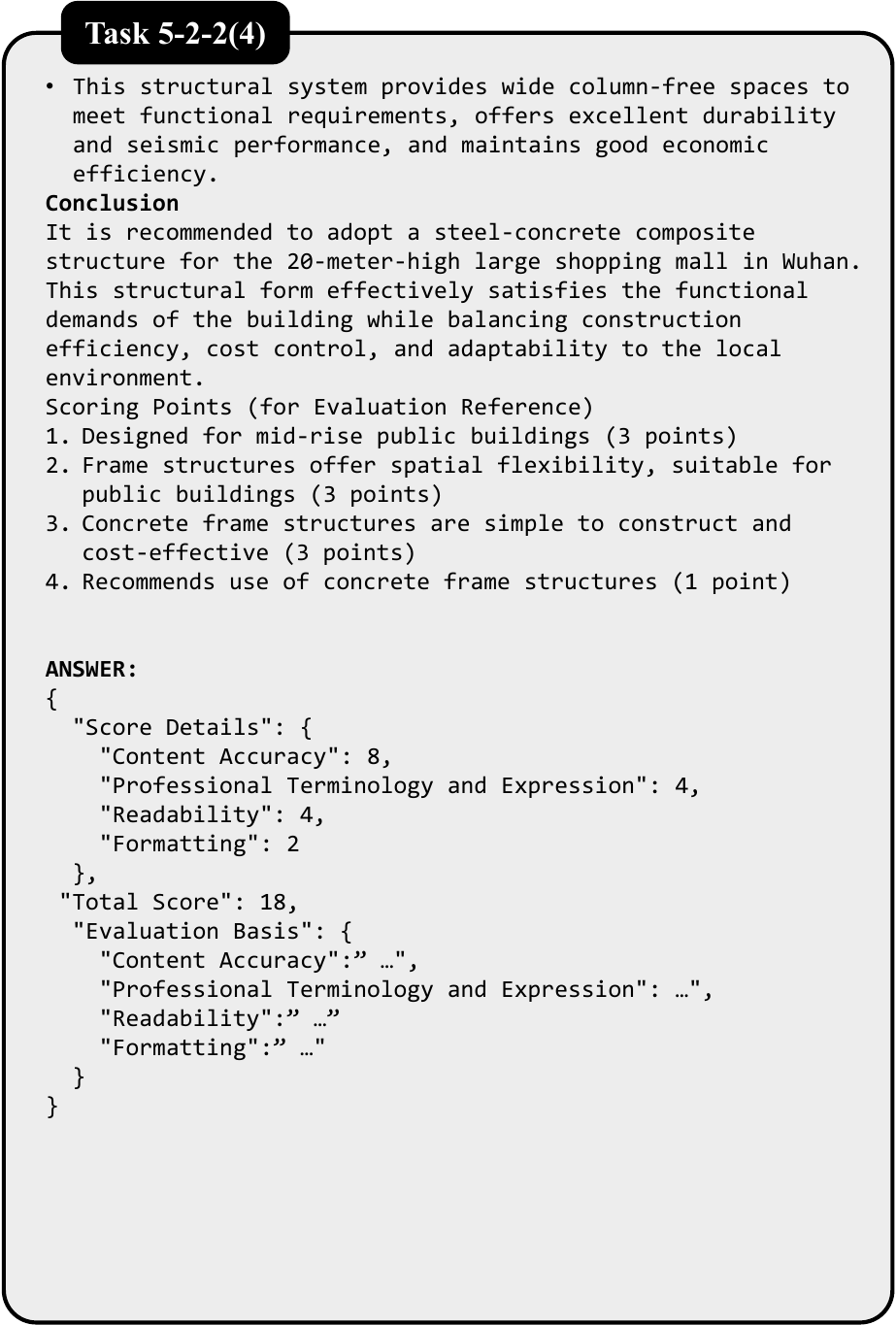}    \caption*{Figure \thefigure: The instruction and an example of Task 5-2-2 Evaluation of Documents in Structural Design(continued)}
	\label{fig:5-2-2c4}
\end{figure}

\clearpage
\begin{figure}[p]   
	\vspace{-20pt}
	\centering
	\includegraphics[width=0.95\textwidth]{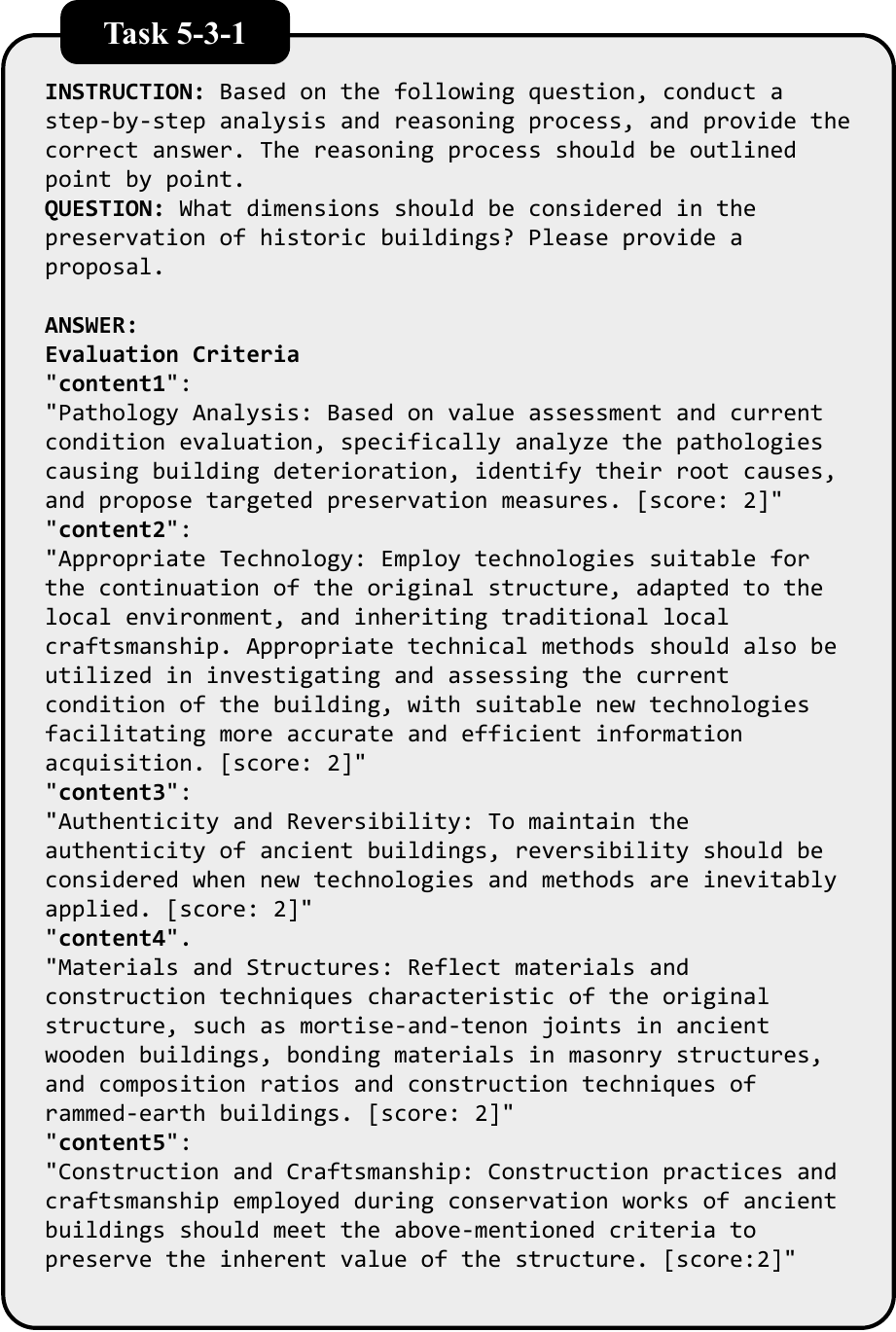}
	\caption{The instruction and an example of Task 5-3-1 Conceptual Design Proposal Generation}
	\label{fig:5-3-1c}
\end{figure}

\clearpage
\begin{figure}[p]   
	\vspace{-20pt}
	\centering
    \includegraphics[width=0.95\textwidth]{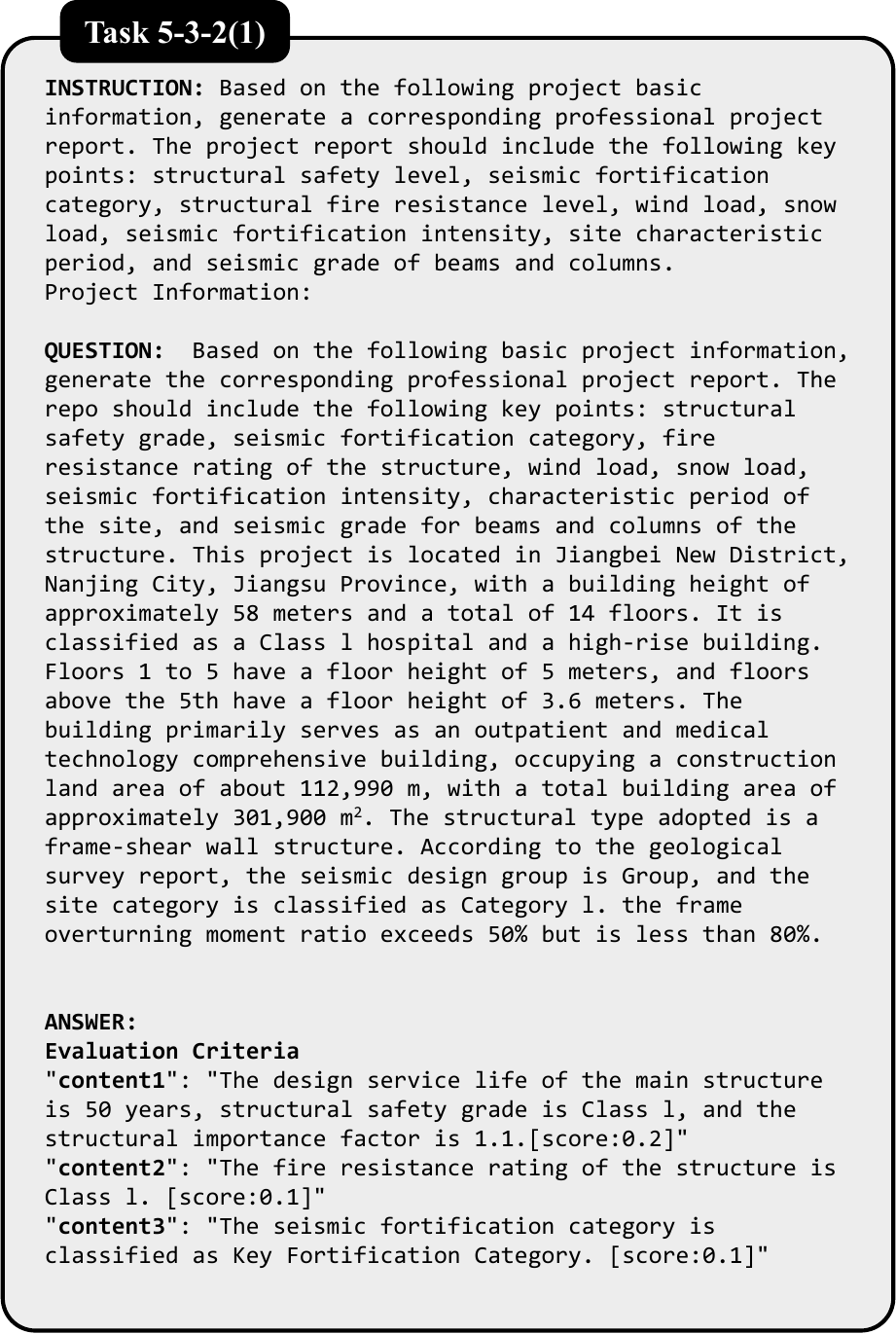}
    \caption{The instruction and an example of Task 5-3-2 Specialized Report Generation\\(continued on the next page)}
    \label{fig:5-3-2c1}
\end{figure}
\clearpage
\begin{figure}[htbp]
	\centering
	\includegraphics[width=0.95\textwidth]{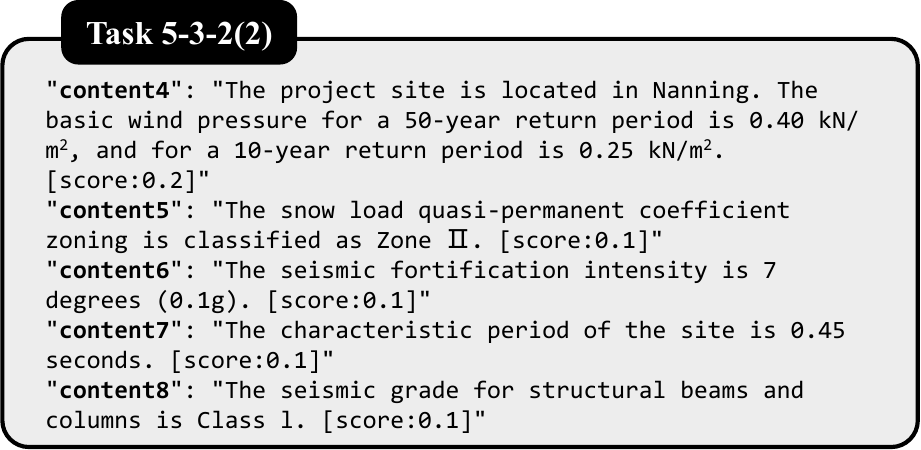}    \caption*{Figure \thefigure: The instruction and an example of Task 5-3-2 Specialized Report Generation(continued)}
	\label{fig:5-3-2c2}
\end{figure}

\clearpage
\begin{figure}[p]   
	\vspace{-20pt}
	\centering
	\includegraphics[width=0.95\textwidth]{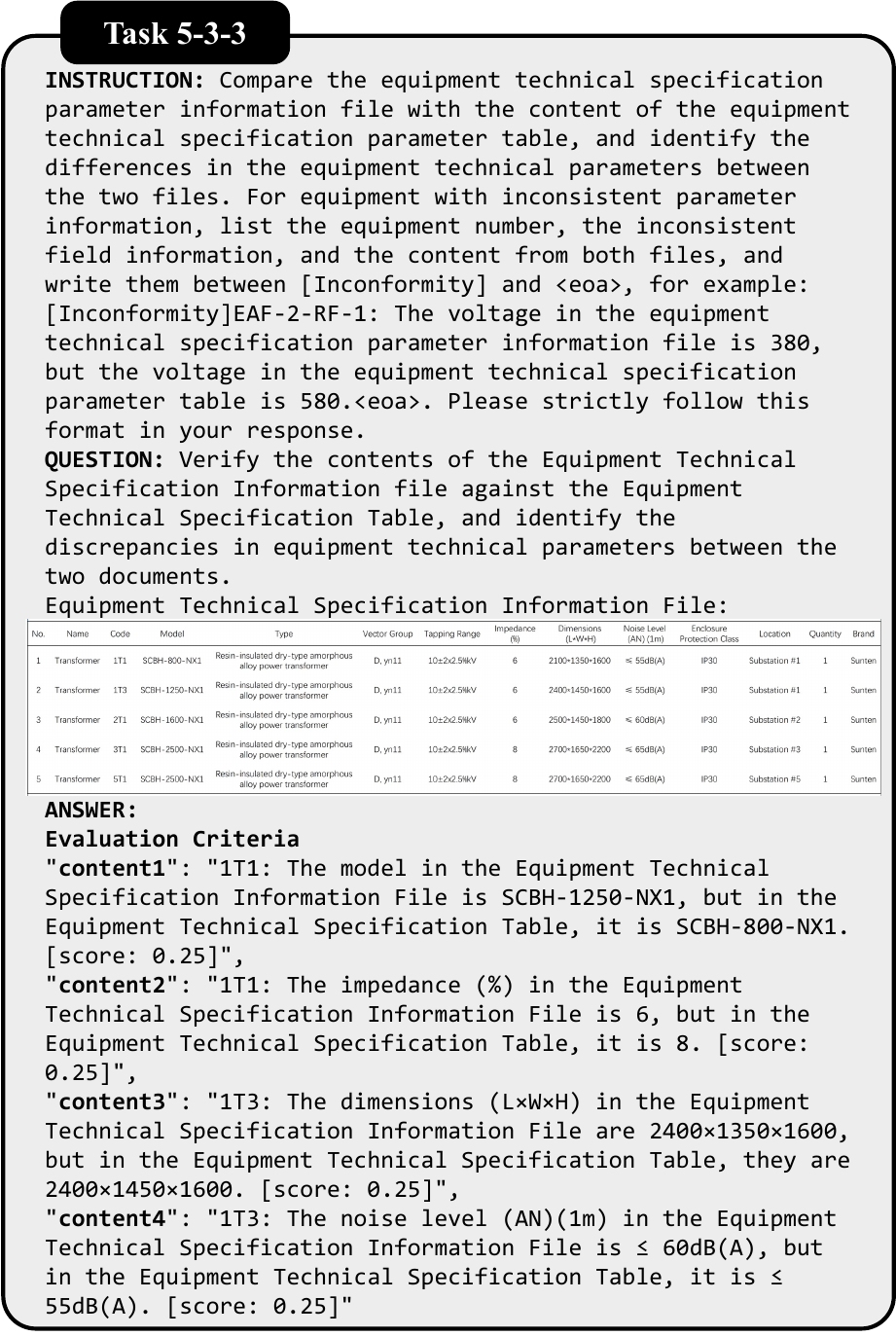}
	\caption{The instruction and an example of Task 5-3-3 Tender Review Report Generation}
	\label{fig:5-3-3c}
\end{figure}

\end{document}